\documentclass[letterpaper, 10 pt, journal, twoside]{IEEEtran}

\IEEEoverridecommandlockouts                              

\usepackage{amsmath} 
\usepackage{amssymb}  

\usepackage{graphicx}
\usepackage{url}
\usepackage{multirow}

\newcommand{\figref}[1]{{Fig.~\ref{#1}}}

\newcommand{\bm}[1]{\mbox{\boldmath{$#1$}}}

\title{\LARGE \bf
  Humanoid Loco-manipulation Planning \\ based on Graph Search and Reachability Maps
}

\author{Masaki Murooka, Iori Kumagai, Mitsuharu Morisawa, Fumio Kanehiro, and Abderrahmane Kheddar
  \thanks{Manuscript received: October, 12, 2020; Revised January, 6, 2021; Accepted February, 3, 2021.}
  \thanks{This paper was recommended for publication by Editor J. Yi upon evaluation of the Associate Editor and Reviewers' comments.} 
  \thanks{The authors are with
    CNRS-AIST JRL (Joint Robotics Laboratory), IRL and
    National Institute of Advanced Industrial Science and Technology (AIST),
    1-1-1 Umezono, Tsukuba, Ibaraki 305-8560, Japan.
    {\tt\footnotesize \{m-murooka, iori-kumagai, m.morisawa, f-kanehiro\}@aist.go.jp, kheddar@gmail.com}}%
  \thanks{Digital Object Identifier (DOI): see top of this page.}
}

\begin{document}

\maketitle

\setlength{\floatsep}{10pt}
\setlength{\textfloatsep}{10pt}
\setlength{\abovecaptionskip}{6pt}

\begin{abstract}
  In this letter, we propose an efficient and highly versatile loco-manipulation planning for humanoid robots.
  Loco-manipulation planning is a key technological brick
  enabling humanoid robots to autonomously perform object transportation by manipulating them.
  We formulate planning of the alternation and sequencing of footsteps and grasps
  as a graph search problem with a new transition model that allows for a flexible representation of loco-manipulation.
  Our transition model is quickly evaluated by relocating and switching the reachability maps
  depending on the motion of both the robot and object.
  We evaluate our approach by applying it to loco-manipulation use-cases,
  such as a bobbin rolling operation with regrasping,
  where the motion is automatically planned by our framework.
\end{abstract}


\section{Introduction}

\IEEEPARstart{M}{oving} large objects is a typical task required for humanoid robots in large-scale manufacturing environments.
As most of such objects are heavy,
they need to be moved through manipulating them by taking advantage of the ground and any possible inertia properties.
Therefore, to accomplish such a task,
robots must autonomously plan loco-manipulation motion;
that is to say, a multi-contact motion that alternates or simultaneously performs bipedal locomotion and object manipulation.

In this letter, we propose a versatile planning framework for loco-manipulation that has the following features:
(i)~determining the robot and object motion while considering obstacle avoidance,
(ii)~stepping a foot while moving an object,
(iii)~(re)grasping an object when necessary,
and finally (iv) treating objects in sliding or rolling motion in a unified manner.
Several studies of humanoid loco-manipulation exist, yet none of them has all of these features integrated~\cite{ManipulationStrategyDecision:Murooka:ICRA2014,PerceptionBasedLocomotion:Kumagai:IROS2018,TeleopPush:Stilman:ICRA2008,Pivot:Yoshida:AR2009,OpenDoor:Nozawa:IROS2012,Locomanip:Jorgensen:ICRA2020,DoorOpen:Dalibard:Humanoids2010,LocomanipNao:Ferrari:ICRA2017,LocomanipWalkman:Settimi:Humanoids2016,MultiModalPlan:Hauser:IJRR2011,SkillMotionPlan:Kallmann:ICRA2010,DualManip:Vahrenkamp:IROS2009}.

\begin{figure}[tpb]
  \begin{center}
    \includegraphics[width=1.0\columnwidth]{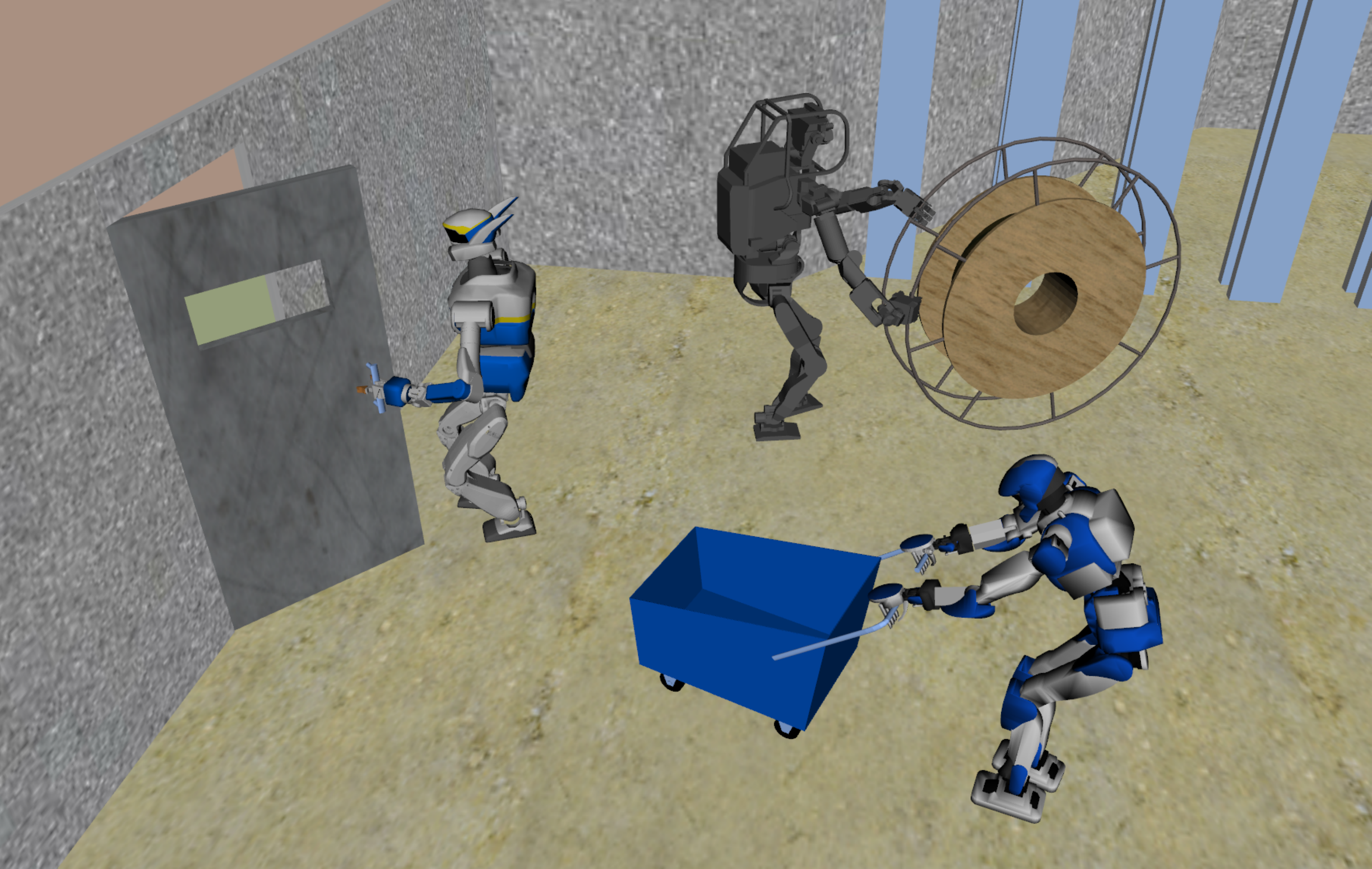}
    \caption{Humanoid loco-manipulation motions.
    }
    \label{fig:intro}
  \end{center}
\end{figure}

The key novelty of our framework is a transition model that
can represent the complicated combinations of locomotion and manipulation primitives.
This transition model can be evaluated efficiently while considering the movement of the grasping point due to the object motion
by switching the reachability maps.
We show that various loco-manipulation behaviors as shown in~\figref{fig:intro} are generated quickly and flexibly
by applying this transition model to a sophisticated graph search algorithm.
It is noteworthy that, to the best of our knowledge,
this is the first study in which
a humanoid robot automatically plans the motion of moving a large cylindrical object by the rolling operation with regrasping.

\subsection{Related Works}

\subsubsection{Large Object Manipulation}

Various types of large object manipulation are achieved by life-size humanoid robots~\cite{ManipulationStrategyDecision:Murooka:ICRA2014}.
When an object is lifted and carried, only the bipedal footsteps are usually planned as the object can be moved in an arbitrary direction~\cite{PerceptionBasedLocomotion:Kumagai:IROS2018}.
In non-prehensile manipulation such as pushing~\cite{TeleopPush:Stilman:ICRA2008} and pivoting~\cite{Pivot:Yoshida:AR2009}
and articulated environment operation such as door opening~\cite{OpenDoor:Nozawa:IROS2012},
it is necessary to consider the kinematic constraints of the object due to the ground contact or the joint.
In many of these works, given a specific operation, ad-hoc rules (e.g., specifying the robot position with respect to the object)
are applied~\cite{TeleopPush:Stilman:ICRA2008,Pivot:Yoshida:AR2009,OpenDoor:Nozawa:IROS2012},
that cannot deal with the situation where the relative position of the robot and the object need to be changed to avoid obstacles.
Our loco-manipulation planning method can cope with such situations and even with situations where object regrasping is necessary.

\subsubsection{Footstep and Multi-contact Planning}

In humanoid bipedal walking, left and right foot is switched alternately,
so graph search that can handle discrete state transitions is often used~\cite{BipedPlan:Chestnutt:Humanoids2003,FootstepPlan:Perrin:Reference2018}.
In a complicated environment, anytime graph search algorithms~\cite{ARAstar:Likhachev:NIPS2004}
are used that quickly acquire an initial suboptimal solution and gradually converge the solutions to the optimal one~\cite{FootstepPlan:Hornung:Humanoids2012}.
In multi-contact motion, which is a more flexible style of behavior, sampling-based and graph search-based methods are used widely~\cite{HumanoidPlanning:Bouyarmane:AR2012,ContactPlanner:Tonneau:TRO2018}.
As a contrasting approach from these, optimization-based planning of bipedal and multi-contact motion has been proposed recently with the advantage of being able to handle motion without discretization~\cite{TrajectoryOptimization:Posa:IJRR2014,OptimizationAtlas:Kuindersma:AuRo2016}.
However, in this letter, graph search is used to incorporate object motion-dependent reachability maps in a concise and low computational cost manner.
We extend the transition model from the bipedal one~\cite{BipedPlan:Chestnutt:Humanoids2003,FootstepPlan:Perrin:Reference2018} so that motions of both robot and object are considered simultaneously.
By limiting the target locomotion form to bipedal walking, loco-manipulation planning is appropriately divided and an efficient transition model is proposed.

\subsubsection{Loco-manipulation Planning}

The motions covered by the previous works concerning humanoid loco-manipulation can roughly be divided into two categories:
the motions that constrain the trajectory of the hands or the manipulated object~\cite{Locomanip:Jorgensen:ICRA2020,DoorOpen:Dalibard:Humanoids2010}
and those that do not~\cite{LocomanipNao:Ferrari:ICRA2017,LocomanipWalkman:Settimi:Humanoids2016}.
The former motions, which are the focus of this paper, are required for
tasks in which the object motion is explicitly constrained, such as opening a door and pushing an object along a trajectory,
while the latter motions are useful for approaching and lifting objects.
Sampling-based multi-modal planning has been proposed for motion that combines stepping and reaching \cite{MultiModalPlan:Hauser:IJRR2011,SkillMotionPlan:Kallmann:ICRA2010}.
The motion along the given hand trajectory is planned
by searching the footstep and object progression at the same time
based on the loco-manipulation reachability acquired by data-driven learning~\cite{Locomanip:Jorgensen:ICRA2020}.
Our approach is similar to this;
however, by relocating and switching the pre-generated reachability maps~\cite{DualManip:Vahrenkamp:IROS2009},
we propose a more general and faster method that can handle regrasping and object rolling.

\subsection{Contributions of this Letter}

The contributions of our work are as follows:
(i) a general loco-manipulation planning method using a flexible transition model with low evaluation cost,
(ii) automatically generating a rolling operation with regrasping by switching the reachability map according to the object motion,
(iii) showing that humanoid motions for various loco-manipulation tasks can be planned in a unified planning framework.

\section{Method Overview} \label{sec:method-overview}

\begin{figure}[tpb]
  \begin{center}
    \includegraphics[width=1.0\columnwidth]{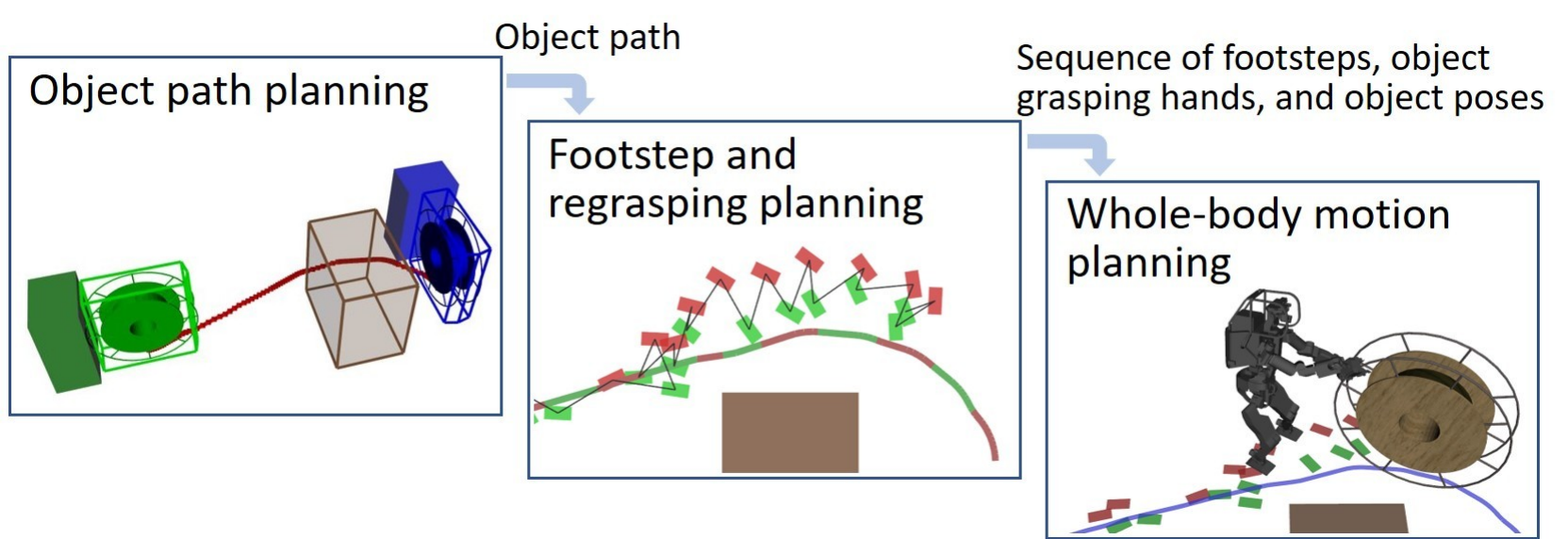}
    \caption{Overview of the planning framework.
      \newline \footnotesize
      The loco-manipulation planning framework consists of three processes:
      object path planning (OP-planning),
      footstep and regrasping planning (FR-planning),
      and whole-body motion planning (WBM-planning).
    }
    \label{fig:process}
  \end{center}
\end{figure}

To make loco-manipulation planning easier to treat,
we divide it into three processes as shown in \figref{fig:process}:
object path planning (OP-planning),
footstep and regrasping planning (FR-planning),
and whole-body motion planning (WBM-planning).

In the following, OP-planning, FR-planning, and WBM-planning are presented in Sections~\ref{sec:op-planning}, \ref{sec:fr-planning}, and \ref{sec:wbm-planning}, respectively.
Especially, Section~\ref{sec:fr-planning} provides a detailed explanation of the transition model that is the main novelty of this letter.
Section~\ref{sec:application} shows the application to various loco-manipulation tasks.

\section{Object Path Planning} \label{sec:op-planning}

\begin{figure}[tpb]
  \begin{center}
    \includegraphics[width=1.0\columnwidth]{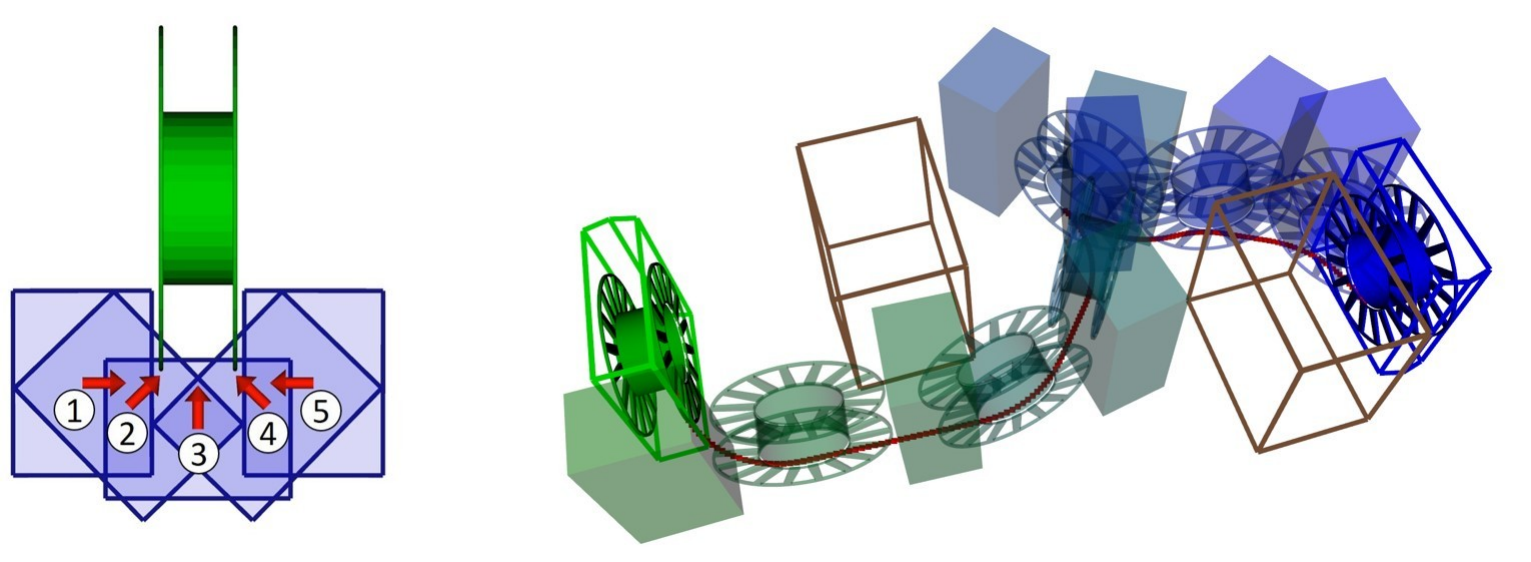}
    \caption{An example of models and results in OP-planning.
      \newline \footnotesize
      The shape of the robot is approximated by a bounding box.
      Five candidates for the robot (in its bounding box) pose relative to the object are shown as an example.
      The robot pose is discretized and represented as the index of the candidates.
    }
    \label{fig:op-plan}
  \end{center}
\end{figure}

We use the asymptotically-optimal sampling-based method RRT*~\cite{RRTstar:Karaman:IJRR2011}
to plan an object path from the start to the goal.
The planning state space is a compound space consisting of the object pose and the robot pose
because it is necessary to consider collision avoidance with the environment not only for the object but also for the robot.
The object pose is represented as the $\mathit{SE}(2)$ space state.
For objects with car-like nonholonomic motion constraints,
the Reeds-Shepp curve~\cite{ReedsSheppCurve:Reeds:PJM1990} is used for the edges connecting the states.
To reduce the dimensions of the state space,
the robot pose is represented as the index of the pose candidates relative to the object
(\figref{fig:op-plan}).

\section{Footstep and Regrasping Planning} \label{sec:fr-planning}

\subsection{Problem Settings} \label{sec:fr-planning-settings}

Footstep and regrasping planning (FR-planning) for loco-manipulation
is defined as the problem of finding a sequence of footsteps and object poses
that makes a transition from the initial state to the goal state.
The formulation is made to allow
the transition of regrasping the object from one hand to the other one.

\begin{figure}[tpb]
  \begin{center}
    \includegraphics[width=1.0\columnwidth]{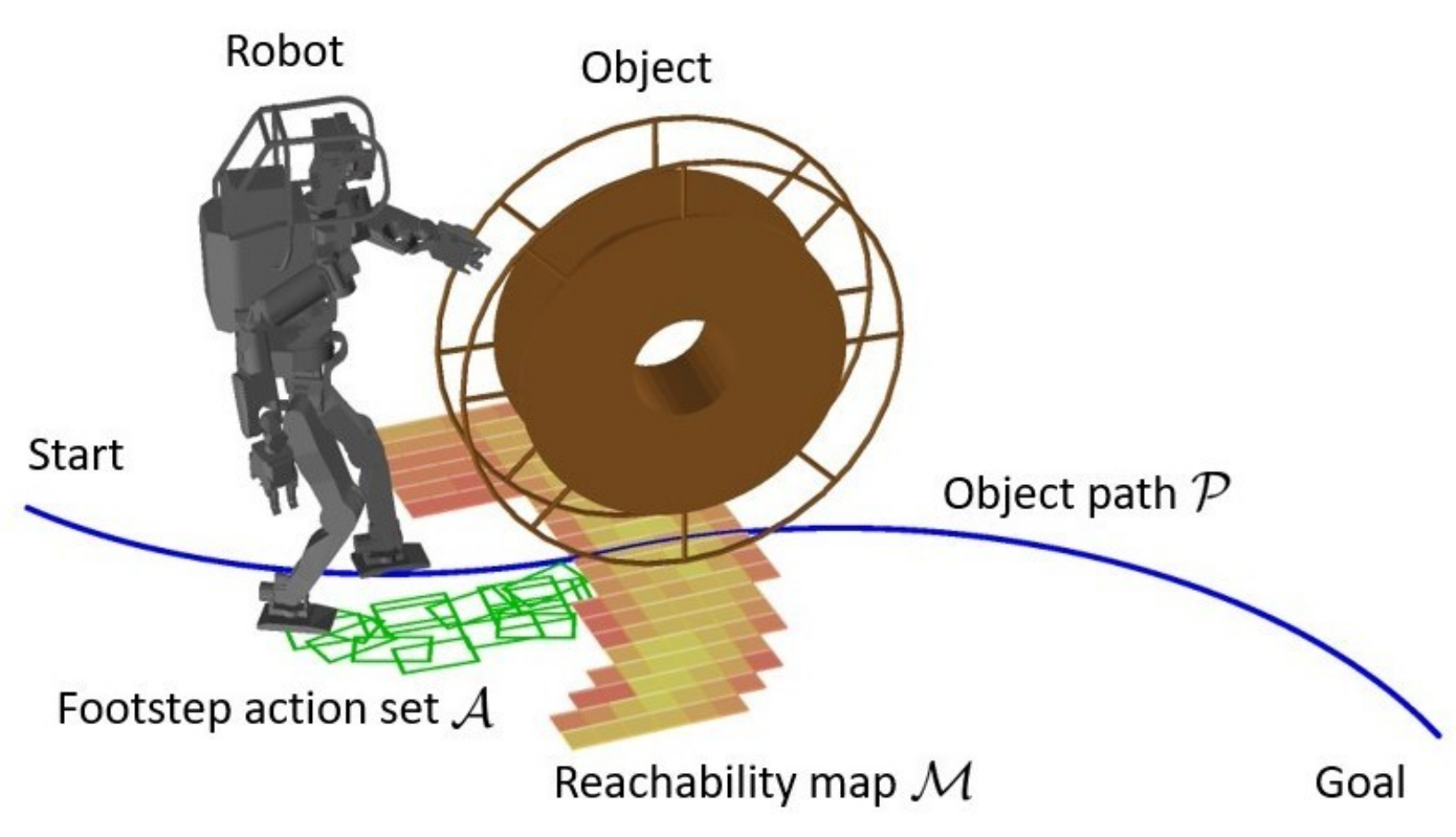}
    \caption{Problem settings of FR-planning.
      \newline \footnotesize
      A robot is moving an object from the start position to the goal position.
      Green rectangle markers represent the footstep action set for the right foot;
      a blue line represents the object path;
      and markers with yellow to red gradation represent the reachability map.
    }
    \label{fig:problem-definition}
  \end{center}
\end{figure}

The following information (\figref{fig:problem-definition}) about the robot and object is assumed to be known in FR-planning:
\begin{itemize}
\item $\mathcal{A}(l_{\mathit{foot}}) =
  \left\{\, \bm{a}{[}i{]} \,\middle|\, i = 1, 2, \cdots, N_\mathcal{A} \,\right\}
  \, (l_{\mathit{foot}} \in \{ L, R \})$:
  a set of discretized footstep actions.
  $N_\mathcal{A}$ is the number of actions.
  $L$ and $R$ represent the left and right foot, respectively.
  Each action $\bm{a} \in \mathit{SE}(2)$ 
  represents the stepping position relative to the opposite foot.
\item $\mathcal{P} = \left\{\, \bm{c}_{\mathcal{P}}{[}i{]} \,\middle|\, i = 1, 2, \cdots, N_\mathcal{P} \,\right\}$:
  a sequence of the object poses that represents the discretized object path.
  Object pose $\bm{c}_{\mathcal{P}} \in \mathit{SE}(2)$ is represented by projecting it onto the ground plane.
  $N_\mathcal{P}$ is the sequence length.
\item $\mathcal{M}(\bm{c}_{\mathit{com}}, l_{\mathit{hand}}) \subset \mathit{SE}(2)$:
  a reachability map of the object.
  The elements in a reachability map are the object poses such that
  the hand $l_{\mathit{hand}} \in \{ L, R \}$ can reach the object
  when the CoM frame of the robot is $\bm{c}_{\mathit{com}} \in \mathit{SE}(2)$.
\end{itemize}
The object path $\mathcal{P}$ is obtained from the previous process, OP-planning.
The procedures for determining the footstep action set $\mathcal{A}$ and the reachability map $\mathcal{M}$
are described in Sections \ref{sec:result-evaluation} and \ref{sec:transition-model}, respectively.

\subsection{Formulation as Graph Search Problem}

Since footstep and regrasping cause discrete state transitions,
we formulate FR-planning as a graph path search problem that is suitable for treating discrete transition combinations.
We focus on two improvements in graph search for FR-planning:
(i) anytime approach that gradually improves the solution~\cite{ARAstar:Likhachev:NIPS2004}
and (ii) incremental replanning for graph updates~\cite{Dstar:Koenig:AAAI2002}.
We use the Anytime Dynamic A* (AD*) algorithm~\cite{ADstar:Likhachev:ICAPS2005} that has these two features for FR-planning.
The graph search algorithms introduced above are generalized
so that a solution can be automatically obtained by giving a graph instance that represents the problem to be solved.
Therefore, we do not explain the details of the graph search algorithm itself, see~\cite{ARAstar:Likhachev:NIPS2004,Dstar:Koenig:AAAI2002,ADstar:Likhachev:ICAPS2005} and the source code~\cite{SBPL:github2020}.
We rather focus on graph components specific to FR-planning in the following.

\subsubsection{State}

In FR-planning, each graph node represents the state defined as follows:
\begin{eqnarray}
  \label{eq:state}
  &&\bm{s} = (
  \bm{c}_{\mathit{st\mathchar`-foot}}, l_{\mathit{st\mathchar`-foot}},
  \bm{c}_{\mathit{sw\mathchar`-foot}}, l_{\mathit{sw\mathchar`-foot}},
  \bm{c}_{\mathit{obj}}, l_{\mathit{hand}}) \ \ \ \ \\
  &&{\rm where} \ \ \bm{c}_{\mathit{st\mathchar`-foot}}, \bm{c}_{\mathit{sw\mathchar`-foot}}, \bm{c}_{\mathit{obj}}
  \in \mathit{SE}(2) \nonumber\\
  &&\phantom{\rm where} \ \ l_{\mathit{st\mathchar`-foot}}, l_{\mathit{sw\mathchar`-foot}}, l_{\mathit{hand}} \in \{L, R\} \nonumber\\
  &&\phantom{\rm where} \ \ l_{\mathit{sw\mathchar`-foot}} = \bar{l}_{\mathit{st\mathchar`-foot}} \nonumber
\end{eqnarray}
where
$\bm{c}_{*}$ is the pose of the foot and object, and $l_{*}$ is the left/right label of the foot and the hand.
$\mathit{st\mathchar`-foot}$ and $\mathit{sw\mathchar`-foot}$ represent a stance foot and a swing foot, respectively.
$\mathit{hand}$ represents a hand that grasps an object.
$\bar{l}_{*}$ represents the opposite label (i.e., $\bar{L}=R, \bar{R}=L$).
Here, we assume that the object is grasped and manipulated with one hand except during regrasping, see Section~\ref{sec:planning-result} for both hands case.

This state definition is extended from the bipedal footstep planning
where $\bm{s}$ is
$(\bm{c}_{\mathit{sw\mathchar`-foot}}, l_{\mathit{sw\mathchar`-foot}})$~\cite{BipedPlan:Chestnutt:Humanoids2003,FootstepPlan:Perrin:Reference2018}
and the state-of-the-art loco-manipulation planning
where $\bm{s}$ is
$(\bm{c}_{\mathit{st\mathchar`-foot}}, l_{\mathit{st\mathchar`-foot}}, \bm{c}_{\mathit{sw\mathchar`-foot}}, l_{\mathit{sw\mathchar`-foot}}, \bm{c}_{\mathit{obj}})$
($l_{\mathit{hand}}$ is missing)~\cite{Locomanip:Jorgensen:ICRA2020}.
The state includes the stance foot, which is unnecessary for bipedal footstep planning,
to evaluate the state transition feasibility as described in Section \ref{sec:transition-model}.

\subsubsection{Successors}

The condition that a state $\bm{s}^{[k+1]}$ is a successor of the state $\bm{s}^{[k]}$
is represented as follows:
\begin{subequations}
  \label{eq:successors}
  \begin{eqnarray}
    &&\hspace{-8mm}
    l_{\mathit{st\mathchar`-foot}}^{[k+1]} = l_{\mathit{sw\mathchar`-foot}}^{[k]}
    \label{eq:successors-st-label} \\
    &&\hspace{-8mm}
    l_{\mathit{sw\mathchar`-foot}}^{[k+1]} = l_{\mathit{st\mathchar`-foot}}^{[k]}
    \label{eq:successors-sw-label} \\
    &&\hspace{-8mm}
    l_{\mathit{hand}}^{[k+1]} \in \{L, R\}
    \label{eq:successors-hand-label} \\
    &&\hspace{-8mm}
    \bm{c}_{\mathit{st\mathchar`-foot}}^{[k+1]} =
    \bm{c}_{\mathit{sw\mathchar`-foot}}^{[k]}
    \label{eq:successors-st-pose} \\
    &&\hspace{-8mm}
    \bm{c}_{\mathit{sw\mathchar`-foot}}^{[k+1]} \in
    \left\{ F_{\mathit{apply}}(\bm{c}_{\mathit{st\mathchar`-foot}}^{[k+1]}, \bm{a})
    \,\middle|\, \bm{a} \in \mathcal{A}(l_{\mathit{sw\mathchar`-foot}}^{[k+1]}) \right\}
    \label{eq:successors-sw-pose} \\
    &&\hspace{-8mm}
    \bm{c}_{\mathit{obj}}^{[k+1]} \in
    \left\{ \bm{c}_{\mathcal{P}}\left[ \mathit{idx}(\bm{c}_{\mathit{obj}}^{[k]}) + i \right] \,\middle|\,
    i = 0, 1, \cdots, N_{\mathit{obj}} \right\}
    \label{eq:successors-hand-pose} \\
    &&\hspace{-8mm}
    F_{\mathit{switchable}}(\ldots) = \mathit{true}
    \label{eq:successors-sw-cond} \\
    &&\hspace{-8mm}
    F_{\mathit{movable}}(\ldots) = \mathit{true}
    \label{eq:successors-mv-cond} \\
    &&\hspace{-8mm}
    F_{\mathit{no\mathchar`-collision}}(\ldots) = \mathit{true}
    \label{eq:successors-col-cond}
  \end{eqnarray}
\end{subequations}
The superscript indicates the index of the state and its elements
(e.g., $l_{\mathit{st\mathchar`-foot}}^{[k]}$ is an element of  $\bm{s}^{[k]}$).
The arguments of the functions in
\eqref{eq:successors-sw-cond}-\eqref{eq:successors-col-cond}
are omitted.

The first three expressions~\eqref{eq:successors-st-label}-\eqref{eq:successors-hand-label} are
the transitions of the left/right label of the foot and the hand.
For the foot, the labels of the stance foot and the swing foot are interchanged,
and for the hand, the transition to both the left and right is allowed.
When $l_{\mathit{hand}}^{[k]}$ and $l_{\mathit{hand}}^{[k+1]}$ are the same,
it means to keep grasping the object with the same hand;
otherwise it means regrasping from one hand to the other one.

\begin{figure}[tpb]
  \begin{center}
    \includegraphics[width=0.49\columnwidth]{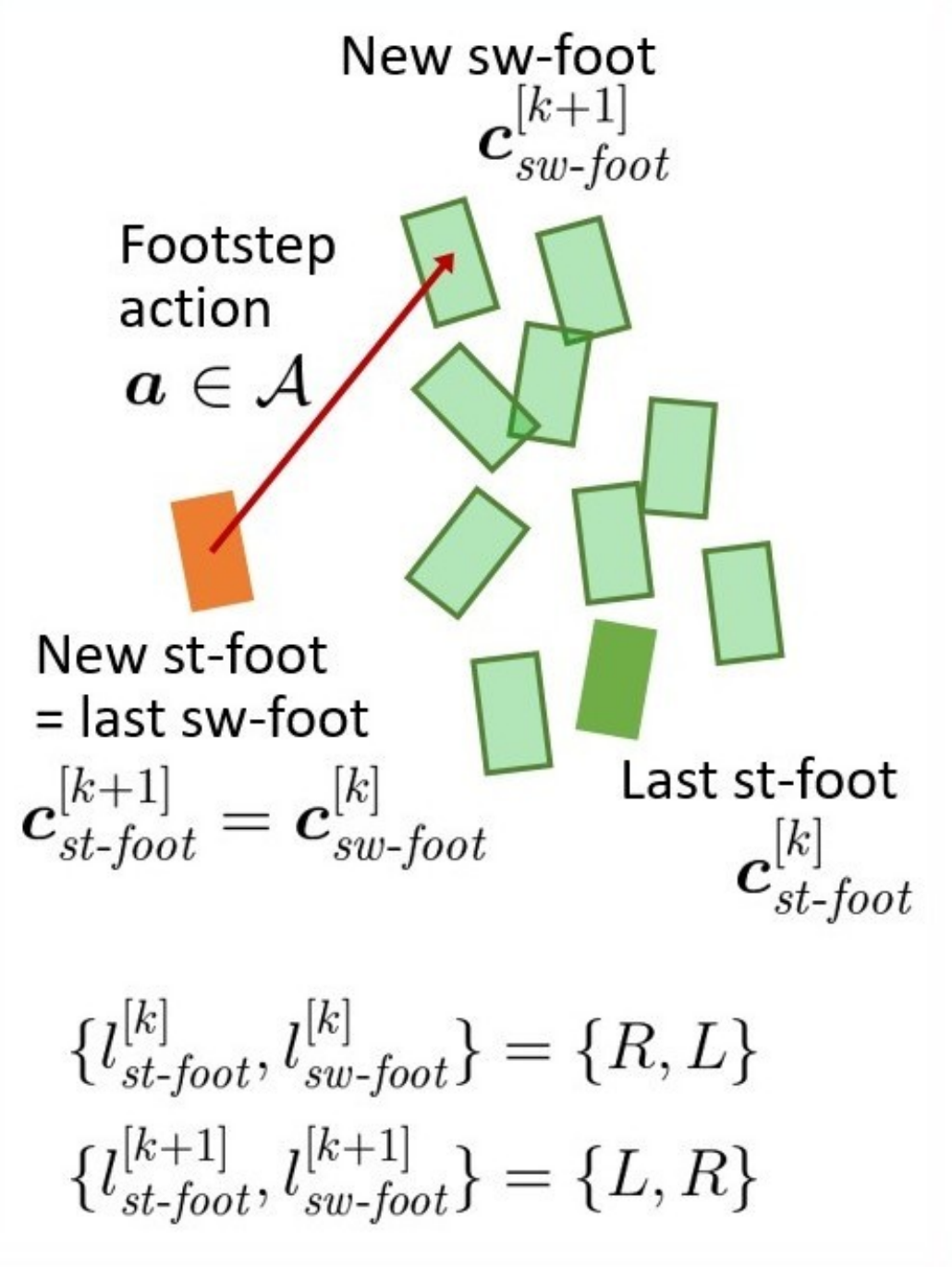}
    \includegraphics[width=0.49\columnwidth]{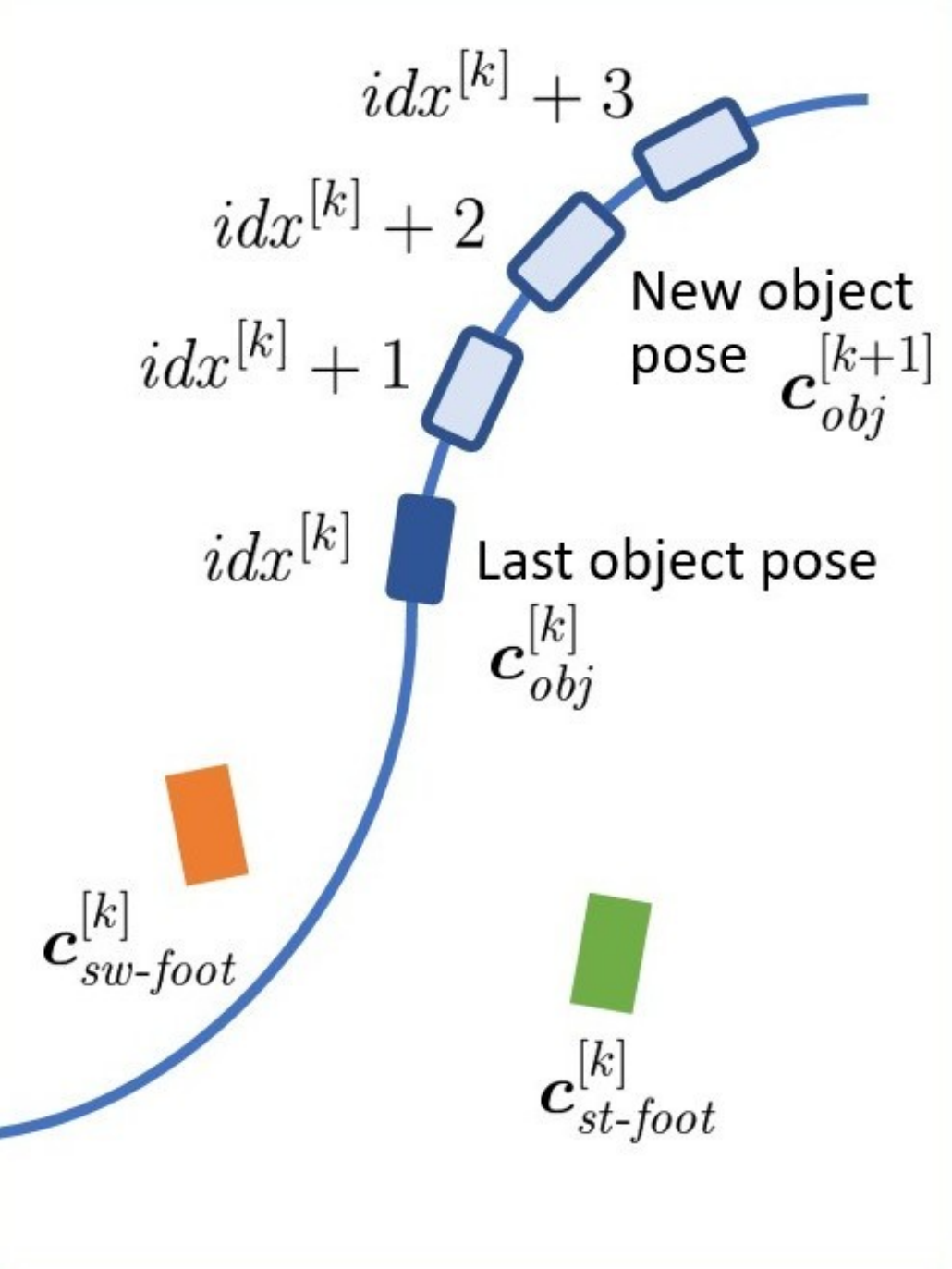} \\
    \begin{minipage}{0.50\columnwidth}
      \begin{center} \footnotesize (A) Footstep transition
      \eqref{eq:successors-st-pose} and \eqref{eq:successors-sw-pose}. \end{center}
    \end{minipage}
    \begin{minipage}{0.48\columnwidth}
      \begin{center} \footnotesize (B) Object pose transition
      \eqref{eq:successors-hand-pose}. \end{center}
    \end{minipage}
    \caption{Transitions in state successors of FR-planning.
      \newline \footnotesize
      Orange markers represent the left foot;
      green markers represent the right foot;
      and blue markers represent the object.
      $\mathit{idx}^{[k]}$ is an abbreviation for $\mathit{idx}(\bm{c}_{\mathit{obj}}^{[k]})$.
    }
    \label{fig:successors}
  \end{center}
\end{figure}

Expressions~\eqref{eq:successors-st-pose} and~\eqref{eq:successors-sw-pose} are
the transitions of the foot poses shown in \figref{fig:successors}~(A).
Since the last swing foot becomes the new stance foot,
the pose of the new stance foot ($\bm{c}_{\mathit{st\mathchar`-foot}}^{[k+1]}$) is the same as the landing pose of the last swing foot ($\bm{c}_{\mathit{sw\mathchar`-foot}}^{[k]}$).
The landing pose of the new swing foot ($\bm{c}_{\mathit{sw\mathchar`-foot}}^{[k+1]}$) is generated
from the footstep action set $\mathcal{A}$.
$F_{\mathit{apply}}(\bm{c}_{\mathit{st\mathchar`-foot}}, \bm{a})$ returns
the swing foot pose after applying the footstep action $\bm{a}$
when the pose of the opposite foot (i.e., the stance foot) is $\bm{c}_{\mathit{st\mathchar`-foot}}$;
recall that the footstep action represents the stepping position relative to the opposite foot.

Expression~\eqref{eq:successors-hand-pose} is
the transition of the object pose shown in \figref{fig:successors}~(B).
The object path $\mathcal{P}$ is discretized, and the index is managed so that the object moves toward the goal by increasing it.
According to the increment from the last index,
successors representing the corresponding object pose are generated.
$\mathit{idx}(\bm{c}_{\mathit{obj}})$ represents an index $i$ such that $\bm{c}_{\mathcal{P}}[i] = \bm{c}_{\mathit{obj}}$.
$N_{\mathit{obj}}$ is the incremental limit of the index in one transition.

Expression~\eqref{eq:successors-sw-cond} is the condition for switching hands for regrasping,
and expression~\eqref{eq:successors-mv-cond} is the condition for moving the object along the path.
The detailed process for evaluating $F_{\mathit{switchable}}$ and $F_{\mathit{movable}}$
based on a reachability map is described in Section \ref{sec:transition-model}.

Expression~\eqref{eq:successors-col-cond} triggers environment's obstacles avoidance.
In FR-planning, geometric collision is detected on the basis of bounding box approximation of shape model.

State~\eqref{eq:state} and state transition~\eqref{eq:successors} are general enough
to flexibly capture loco-manipulation including regrasping.
Yet, pure manipulation can occur since the footstep action set includes no-stepping in~\eqref{eq:successors-sw-pose};
and pure locomotion is possible since zero is included in the index increments of the object pose in~\eqref{eq:successors-hand-pose}.

\subsubsection{Start and Goal}

The start state $\bm{s}^{[0]}$ and goal states $\bm{s}^{[G]}$ are defined as follows:
\begin{subequations}
  \begin{eqnarray}
  &&\bm{c}^{[0]}_{\mathit{obj}} = \bm{c}_{\mathcal{P}}[1] \\
  &&\bm{c}^{[0]}_{\mathit{st\mathchar`-foot}}, \bm{c}^{[0]}_{\mathit{sw\mathchar`-foot}}
  = \mathit{initial \ foot \ poses} \\
  &&\{ l^{[0]}_{\mathit{st\mathchar`-foot}}, l^{[0]}_{\mathit{sw\mathchar`-foot}}, l^{[0]}_{\mathit{hand}} \} = \{ L, R, L \} \\
  &&\bm{c}^{[G]}_{\mathit{obj}} = \bm{c}_{\mathcal{P}}[N_\mathcal{P}]
  \end{eqnarray}
\end{subequations}
The initial foot poses are known.
Left/right labels in the start state are given temporarily, and, if necessary, they will be swapped in successors.
In the goal states, any pose and label are allowed for the foot and the hand.

\subsubsection{State Transition Cost}

The cost to transit from $\bm{s}^{[k]}$ to $\bm{s}^{[k+1]}$ is represented as follows:
\begin{eqnarray}
  c(\bm{s}^{[k]}, \bm{s}^{[k+1]})
  =
  d(\bm{c}_{\mathit{obj}}^{[k]}, \bm{c}_{\mathit{obj}}^{[k+1]})
  + c_{\mathit{step}}
  + c_{\mathit{regrasp}}
\end{eqnarray}
$d(\bm{c}_{\mathit{obj}}^{[k]}, \bm{c}_{\mathit{obj}}^{[k+1]})$ represents the distance along the object path.
$c_{\mathit{step}}$ and $c_{\mathit{regrasp}}$ are the constant costs that are added
only when stepping a foot and regrasping an object, respectively.

\subsubsection{Heuristic Function}

The AD* algorithm efficiently finds a solution by prioritizing searches based on heuristics.
The heuristic function of the state $\bm{s}^{[k]}$ is represented as follows:
\begin{eqnarray}
  h(\bm{s}^{[k]})
  =
  d(\bm{c}_{\mathit{obj}}^{[k]}, \bm{c}_{\mathit{obj}}^{[G]})
  + N_{\mathit{step}} c_{\mathit{step}}
  + h_{\mathit{nominal}}
  \label{eq:heuristics}
\end{eqnarray}
$N_{\mathit{step}}$ is the minimum number of footsteps from the current state to the goal state,
calculated from the maximum stride length of the footstep action set $\mathcal{A}$.
$h_{\mathit{nominal}}$ is a heuristic value according to the distance
between the current foot poses and the nominal foot poses calculated from the current object pose.
Without $h_{\mathit{nominal}}$, the heuristic is admissible~\cite{ARAstar:Likhachev:NIPS2004}.
The details and effectiveness of $h_{\mathit{nominal}}$ are described in Section~\ref{sec:result-evaluation}.

\subsection{Transition Evaluation based on Reachability Map} \label{sec:transition-model}

\subsubsection{Reachability Map Generation}

\begin{figure}[tpb]
  \begin{center}
    \includegraphics[width=0.49\columnwidth]{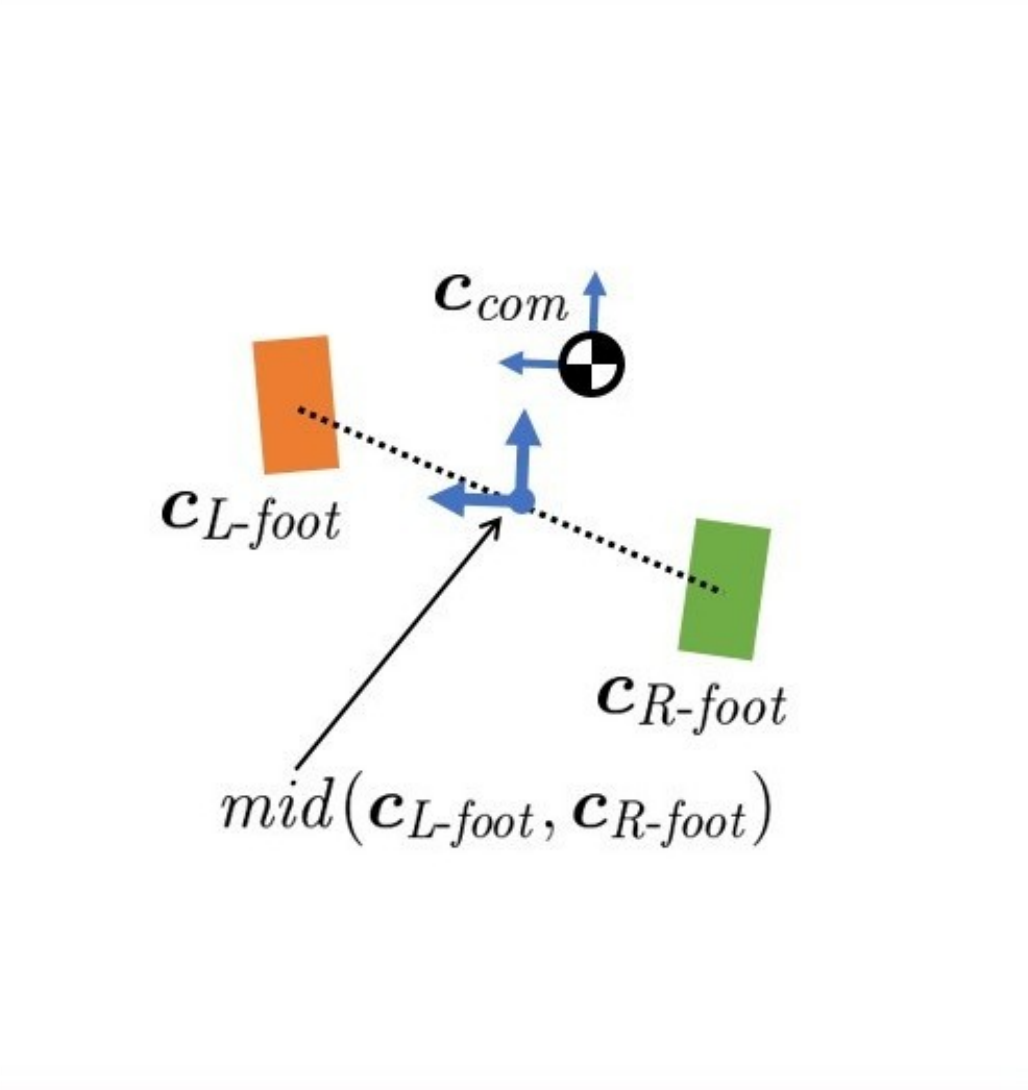}
    \includegraphics[width=0.49\columnwidth]{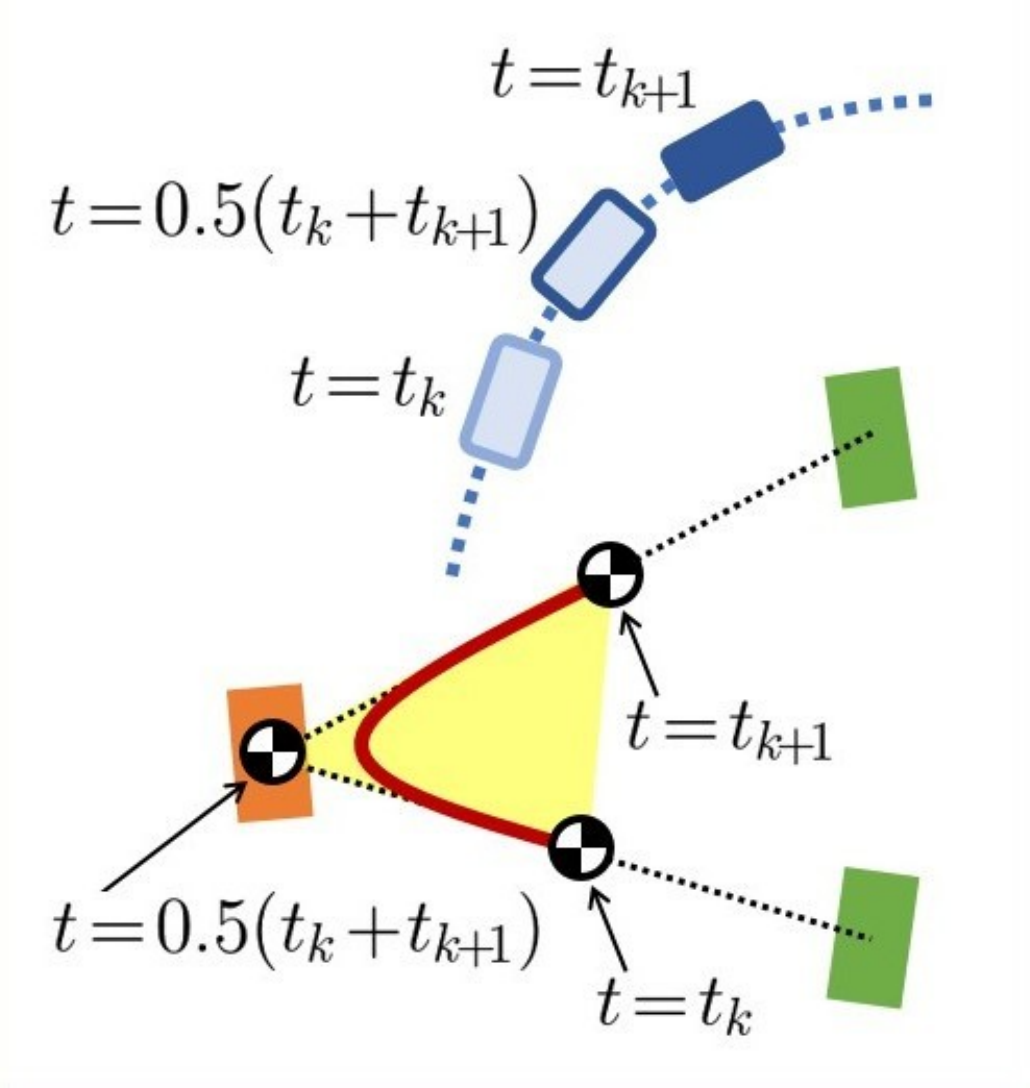} \\
    \begin{minipage}{0.49\columnwidth}
      \begin{center} \footnotesize (A) CoM frame \end{center}
    \end{minipage}
    \begin{minipage}{0.49\columnwidth}
      \begin{center} \footnotesize (B) CoM trajectory \end{center}
    \end{minipage}
    \caption{Frame and trajectory of CoM
      \newline \footnotesize
      (A) $\mathit{mid}(\bm{c}_{\mathit{L\mathchar`-foot}}, \bm{c}_{\mathit{R\mathchar`-foot}})$
      is the middle pose between the left and right foot poses.
      The position of CoM frame $\bm{c}_{\mathit{com}}$ coincides with the robot's CoM,
      and its orientation coincides with the middle pose.
      \newline
      (B) The red curve illustrates an example of the CoM trajectory when stepping the right foot.
      To simplify the movability evaluation,
      the assumptions shown here are imposed
      on the poses of the robot's CoM and the object in the middle timing of the transition.
    }
    \label{fig:trans-com}
  \end{center}
\end{figure}

To avoid repetitive calculation of computationally expensive whole-body inverse kinematics,
a pre-computed reachability map is used for the feasibility evaluation of the transition.
The reachability map $\mathcal{M}(\bm{c}_{\mathit{com}}, l_{\mathit{hand}})$ is represented
as the set of object poses relative to the robot's CoM frame defined as in \figref{fig:trans-com}~(A)
to evaluate the reachability regardless of the foot poses.

\begin{figure}[tpb]
  \begin{center}
    \includegraphics[width=0.49\columnwidth]{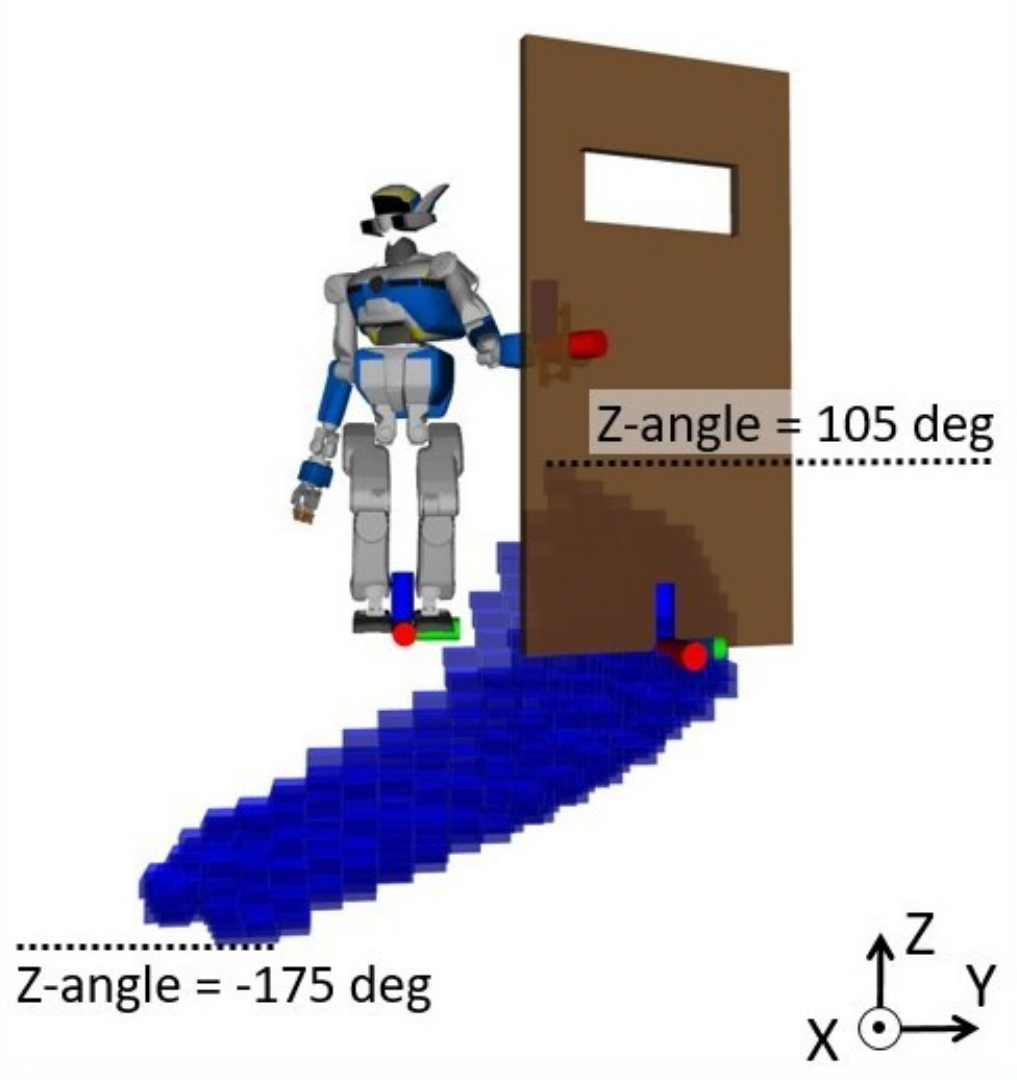}
    \includegraphics[width=0.49\columnwidth]{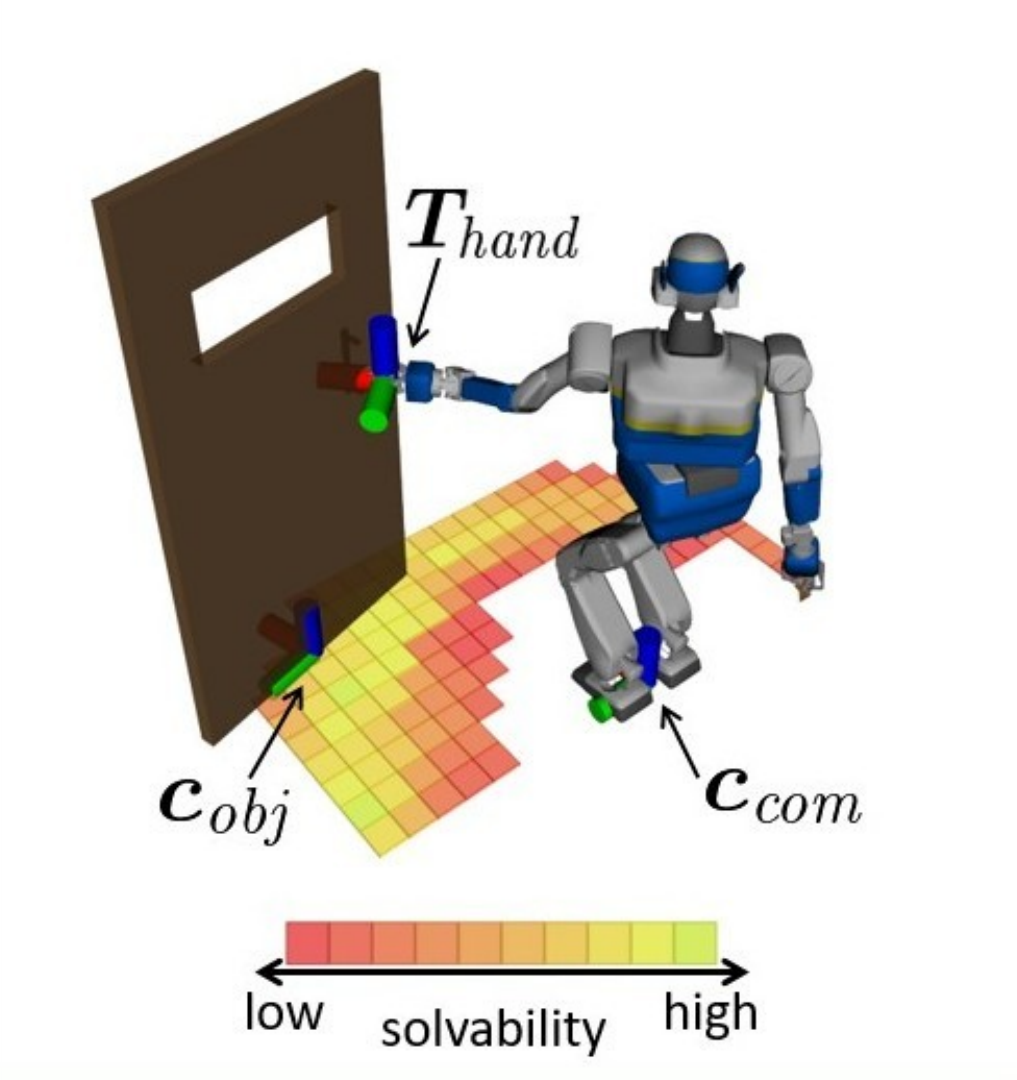} \\
    \begin{minipage}{0.49\columnwidth}
      \begin{center} \footnotesize (A) 3D visualization \end{center}
    \end{minipage}
    \begin{minipage}{0.49\columnwidth}
      \begin{center} \footnotesize (B) 2D visualization \end{center}
    \end{minipage}
    \caption{An example of reachability map.
      \newline \footnotesize
      A reachability map for opening the door by grasping the doorknob with the left hand is shown.
      A reachability map is represented as a set of cells on a 3D grid consisting of the X and Y positions and the Z-angle (i.e., rotation angle around the Z-axis) of an object.
      \newline
      (A) The map is visualized so that the Z-position of the cell corresponds to the Z-angle of the object.
      It can be seen that the reachable Z-angle changes depending on the Y-position of the object.
      \newline
      (B) The reachable Z-angle information of the object is compressed into the cell color.
      The color of each cell represents the solvability (i.e., the number of reachable Z-angle cells) when the object is in the corresponding X and Y positions.
    }
    \label{fig:rmap}
  \end{center}
\end{figure}

\figref{fig:rmap} shows an example of the reachability map.
By dividing the space of the object pose ($\mathit{SE}(2)$) into a grid
and solving the inverse kinematics for each object pose $\bm{c}_{\mathit{obj}}$,
an object region is derived where the robot can reach the hand $l_{\mathit{hand}}$
to the grasping point $\bm{T}_{\mathit{hand}}$ while keeping the CoM frame in the target $\bm{c}_{\mathit{com}}$.
The inverse kinematics calculation is the same as that in the WBM-planning described in Section~\ref{sec:wbm-planning}.
The transformation from the object pose $\bm{c}_{\mathit{obj}}$ to the grasping point $\bm{T}_{\mathit{hand}}$
is treated as known information.

\begin{figure}[tpb]
  \begin{center}
    \includegraphics[width=0.49\columnwidth]{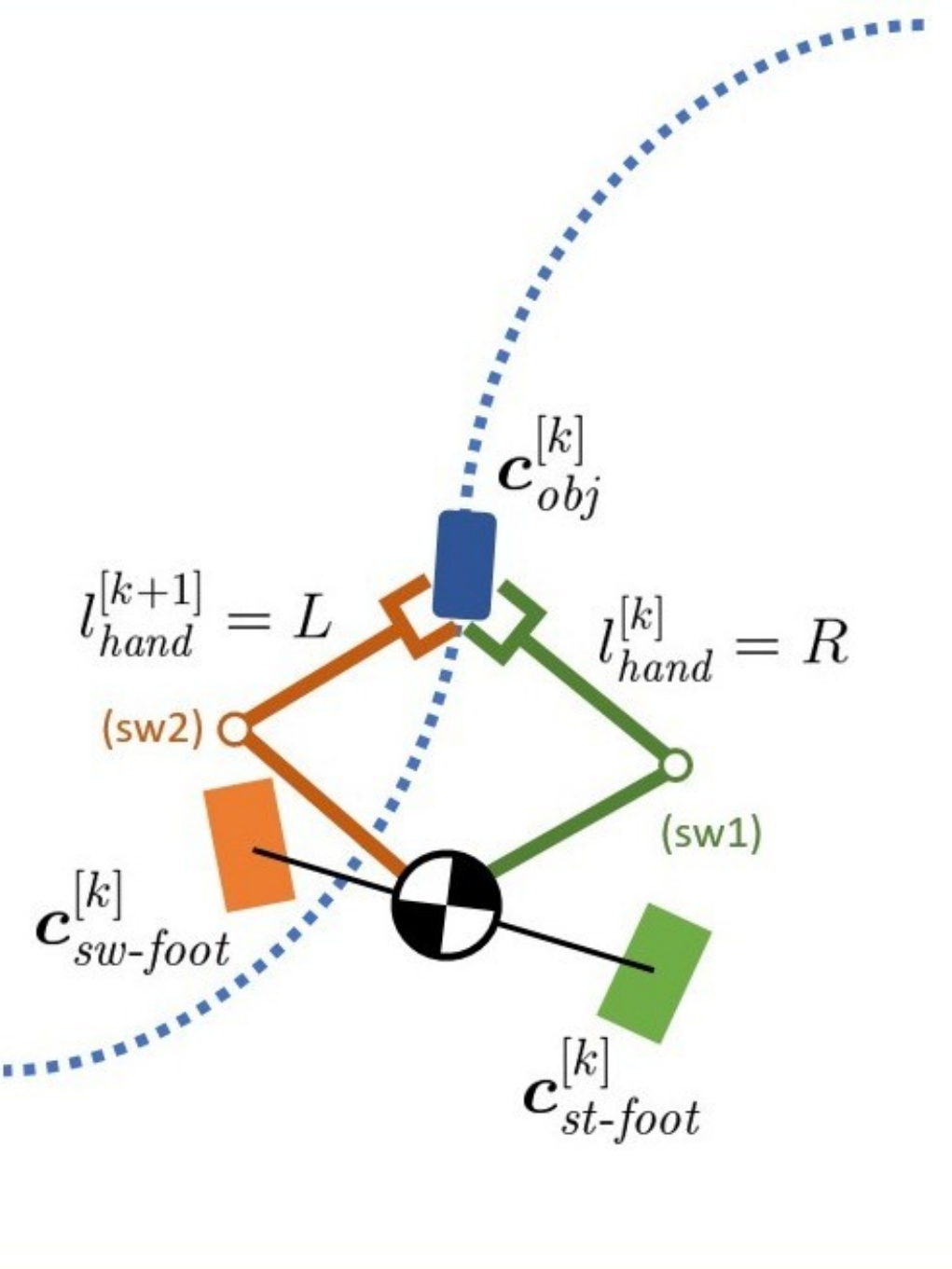}
    \includegraphics[width=0.49\columnwidth]{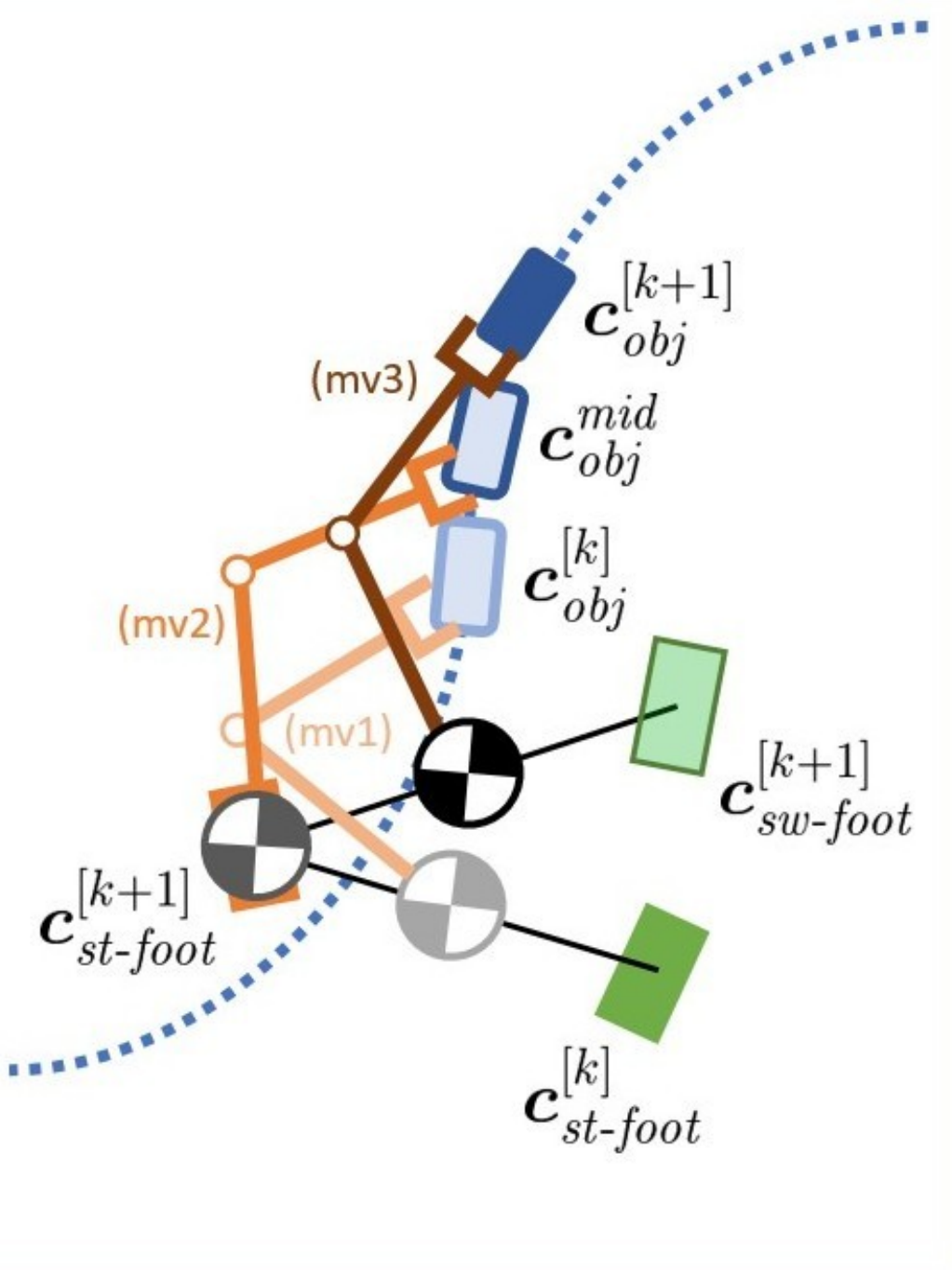} \\
    \begin{minipage}{0.49\columnwidth}
      \begin{center} \footnotesize (A) Switchable condition~\eqref{eq:switchability-eval} \end{center}
    \end{minipage}
    \begin{minipage}{0.49\columnwidth}
      \begin{center} \footnotesize (B) Movable condition~\eqref{eq:movability-eval} \end{center}
    \end{minipage}
    \caption{Transition evaluation based on reachability map.
      \newline \footnotesize
      (A) Switchability when regrasping an object from the right hand to the left hand is evaluated.
      \newline
      (B) Movability of an object with stepping the right foot is evaluated.
      (mv1), (mv2), and (mv3) correspond to
      $t \! = \! t_k$,
      $t \! = \! 0.5 (t_k \! + \! t_{k+1})$, and
      $t \! = \! t_{k+1}$ in \figref{fig:trans-com}~(B), respectively.
    }
    \label{fig:switchable-movable}
  \end{center}
\end{figure}

\subsubsection{Switchability Evaluation}

The switchable condition $F_{\mathit{switchable}}$ in~\eqref{eq:successors-sw-cond} depends on
the last state of the feet and the object
($l_{\mathit{st\mathchar`-foot}}^{[k]}, \bm{c}_{\mathit{st\mathchar`-foot}}^{[k]}, l_{\mathit{sw\mathchar`-foot}}^{[k]}, \bm{c}_{\mathit{sw\mathchar`-foot}}^{[k]}, \bm{c}_{\mathit{obj}}^{[k]}$)
and the transition of hand label
($l_{\mathit{hand}}^{[k]}, l_{\mathit{hand}}^{[k+1]}$).
If $l_{\mathit{hand}}^{[k]}$ and $l_{\mathit{hand}}^{[k+1]}$ are the same, it is always satisfied.
Otherwise,
when both feet and the object are fixed to the last poses
and the hand can be switched from $l_{\mathit{hand}}^{[k]}$ to $l_{\mathit{hand}}^{[k+1]}$,
the switchability is satisfied (\figref{fig:switchable-movable}~(A)).

This condition is formulated using the reachability map as follows:
\begin{eqnarray}
  && \hspace{-8mm}
  F_{\mathit{switchable}}(\ldots) = \nonumber\\
  && \hspace{-4mm}
  \bm{c}_{\mathit{obj}}^{[k]} \in \mathcal{M} \left(
  \mathit{mid}(\bm{c}_{\mathit{st\mathchar`-foot}}^{[k]}, \bm{c}_{\mathit{sw\mathchar`-foot}}^{[k]}),
  l^{[k]}_{\mathit{hand}}
  \right)
  \nonumber \\
  && \hspace{-4mm}
  \land \ \
  \bm{c}_{\mathit{obj}}^{[k]} \in \mathcal{M} \left(
  \mathit{mid}(\bm{c}_{\mathit{st\mathchar`-foot}}^{[k]}, \bm{c}_{\mathit{sw\mathchar`-foot}}^{[k]}),
  l^{[k+1]}_{\mathit{hand}}
  \right)
  \hspace{8mm}
  \label{eq:switchability-eval}
\end{eqnarray}
where $\mathit{mid}(\bm{c}_1, \bm{c}_2)$ represents the middle pose between $\bm{c}_1$ and $\bm{c}_2$.
During regrasping, the robot's CoM is assumed to be on the middle of both feet.
The first evaluation of the reachability map in~\eqref{eq:switchability-eval} corresponds to
(sw1) in \figref{fig:switchable-movable}~(A),
and the second corresponds to (sw2).

\subsubsection{Movability Evaluation}

The movable condition $F_{\mathit{movable}}$ in~\eqref{eq:successors-mv-cond} depends on
the new state of the stance foot and the hand
($l_{\mathit{st\mathchar`-foot}}^{[k+1]}, \bm{c}_{\mathit{st\mathchar`-foot}}^{[k+1]}, l_{\mathit{hand}}^{[k+1]}$),
the transition of the swing foot
($l_{\mathit{sw\mathchar`-foot}}^{[k+1]}, \bm{c}_{\mathit{st\mathchar`-foot}}^{[k]}, \bm{c}_{\mathit{sw\mathchar`-foot}}^{[k+1]}$),
and the transition of the object
($\bm{c}_{\mathit{obj}}^{[k]}, \bm{c}_{\mathit{obj}}^{[k+1]}$).
It is satisfied
when the object can be manipulated
from $\bm{c}_{\mathit{obj}}^{[k]}$ to $\bm{c}_{\mathit{obj}}^{[k+1]}$ by the hand $l_{\mathit{hand}}^{[k+1]}$
while fixing the foot $l_{\mathit{st\mathchar`-foot}}^{[k+1]}$ to $\bm{c}_{\mathit{st\mathchar`-foot}}^{[k+1]}$
and stepping the foot $l_{\mathit{sw\mathchar`-foot}}^{[k+1]}$
from $\bm{c}_{\mathit{st\mathchar`-foot}}^{[k]}$ to $\bm{c}_{\mathit{sw\mathchar`-foot}}^{[k+1]}$
(\figref{fig:switchable-movable}~(B)).

In~\cite{Locomanip:Jorgensen:ICRA2020},
the movable condition is obtained by generating a dataset in the loco-manipulation motion space and training a neural network.
However, due to the large dimensions of the loco-manipulation motion space,
the computational cost of dataset generation is large, even if it is calculated offline.

In this study,
by making an assumption about the motions of the robot's CoM and the object,
the movable condition is derived based on the reachability map evaluations.
\figref{fig:trans-com}~(B) shows these motions during the loco-manipulation transition.
The CoM trajectory depends on the motion properties (e.g., the CoM height and the stepping duration)
and the generation algorithm (e.g.,~\cite{PreviewControl:Kajita:ICRA2003,DcmWalk:Englsberger:TRO2015});
however, in general, it can be approximated as being inside a triangular area connecting the stance foot and the CoM before and after the transition
(visualized in yellow in \figref{fig:trans-com}~(B)).
As the most severe case in terms of reachability,
we assume the CoM is on the stance foot at the middle timing of the transition;
it corresponds to $t \! = \! 0.5 (t_k \! + \! t_{k+1})$ in \figref{fig:trans-com}~(B).
Assuming that the CoM and the object move synchronously,
the object is considered to be in the middle pose of the transition.
Based on these considerations, the movable condition is formulated as follows:
\begin{eqnarray}
  && \hspace{-8mm}
  F_{\mathit{movable}}(\ldots) = \nonumber\\
  && \hspace{-4mm}
  \bm{c}_{\mathit{obj}}^{\mathit{mid}} \in \mathcal{M} \left(
  \bm{c}_{\mathit{st\mathchar`-foot}}^{[k+1]},
  l_{\mathit{hand}}^{[k+1]}
  \right) \nonumber\\
  && \hspace{-4mm}
  \land \ \
  \bm{c}_{\mathit{obj}}^{[k+1]} \in \mathcal{M} \left(
  \mathit{mid}(\bm{c}_{\mathit{st\mathchar`-foot}}^{[k+1]}, \bm{c}_{\mathit{sw\mathchar`-foot}}^{[k+1]}),
  l^{[k+1]}_{\mathit{hand}}
  \right)
  \label{eq:movability-eval} \\
  && \hspace{-8mm}
  {\rm where} \ \
  \bm{c}_{\mathit{obj}}^{\mathit{mid}} =
  \bm{c}_{\mathcal{P}}
  \left[ \left( \mathit{idx}(\bm{c}_{\mathit{obj}}^{[k]}) + \mathit{idx}(\bm{c}_{\mathit{obj}}^{[k+1]}) \right) / \, 2 \right]
  \nonumber
\end{eqnarray}
$\bm{c}_{\mathit{obj}}^{\mathit{mid}}$ represents the object pose in the middle timing of the transition.
The first evaluation of the reachability map in~\eqref{eq:movability-eval} corresponds to
(mv2) in \figref{fig:switchable-movable}~(B),
and the second corresponds to (mv3);
(mv1) has already been checked in the last switchability evaluation.

\subsubsection{Extension of Transition Evaluation to Rolling Objects}

The problem with applying the transition evaluation based on the reachability map to a rolling object is that
the reachability map is not constant because
the transformation from the object pose $\bm{c}_{\mathit{obj}}$ to the grasping point $\bm{T}_{\mathit{hand}}$ changes
according to the motion of the object.
In this study,
this problem is solved by generating multiple reachability maps according to the rolling angles of the object
and switching them according to the moving distance of the object.

\begin{figure}[tpb]
  \begin{center}
    \includegraphics[width=1.0\columnwidth]{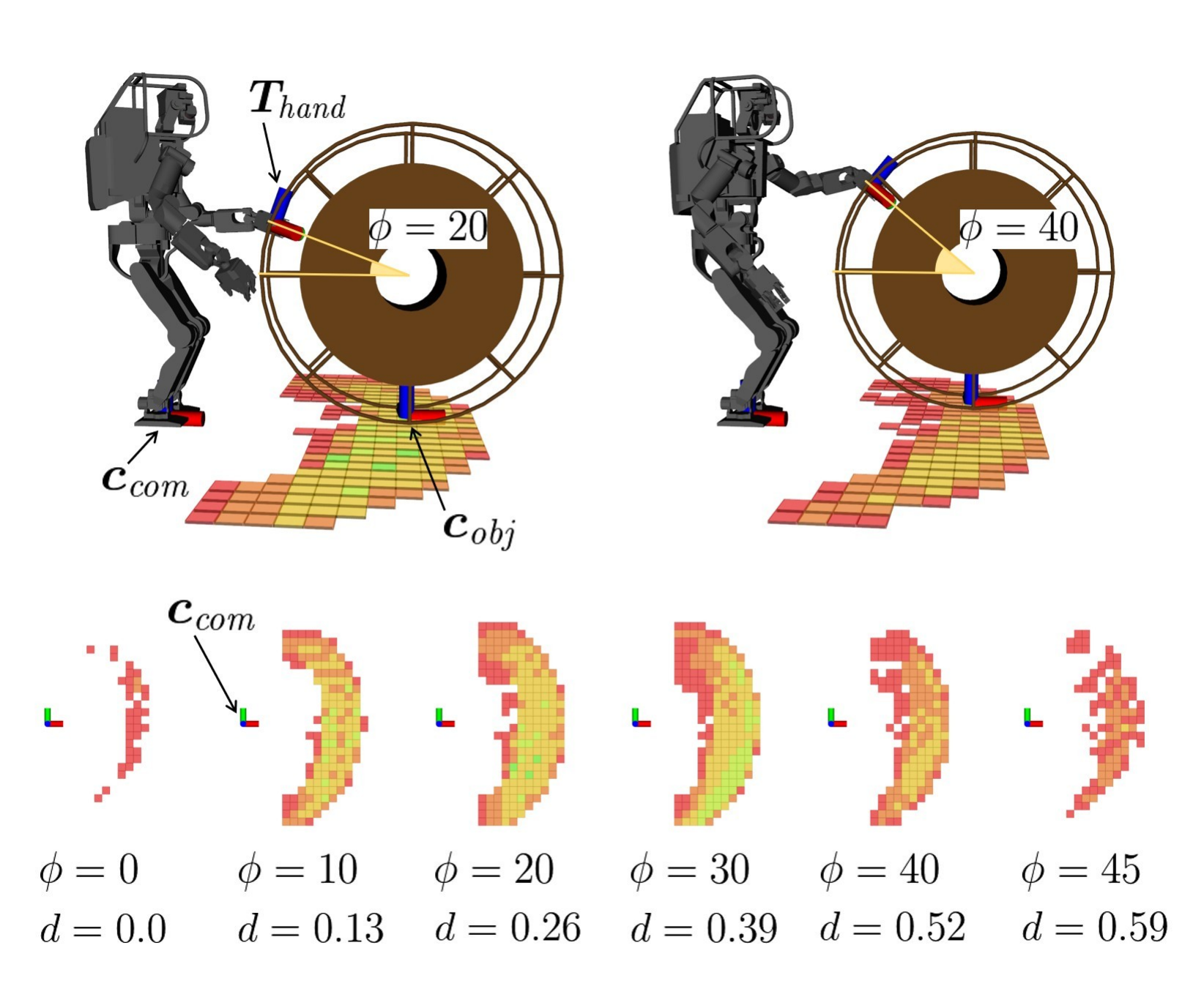}
    \caption{An example of reachability maps for a rolling object.
      \newline \footnotesize
      For each sampled object-rolling angle $\phi$,
      a reachability map is generated for the corresponding grasping point $\bm{T}_{\mathit{hand}}$.
      The object moving distance $d$ is calculated from the rolling angle and the rolling radius.
      It can be seen that the reachable area gradually changes according to the object moving distance.
    }
    \label{fig:rmap-rolling}
  \end{center}
\end{figure}

\figref{fig:rmap-rolling} shows an example of the reachability maps for a rolling object.
The reachable area gradually changes
according to the object moving distance converted from the object-rolling angle.
The set of these maps is represented as $\mathcal{M}(\bm{c}_{\mathit{com}}, l_{\mathit{hand}}, d)$
where $d$ is the distance the object has moved since the last regrasping.
By extending state~\eqref{eq:state} to add an index of the object pose at the last regrasping
and selecting the reachability map according to the distance the object has moved from there,
the FR-planning can be applied to the rolling object.

\section{Whole-body Motion Planning} \label{sec:wbm-planning}

\begin{figure}[tpb]
  \begin{center}
    \includegraphics[width=0.95\columnwidth]{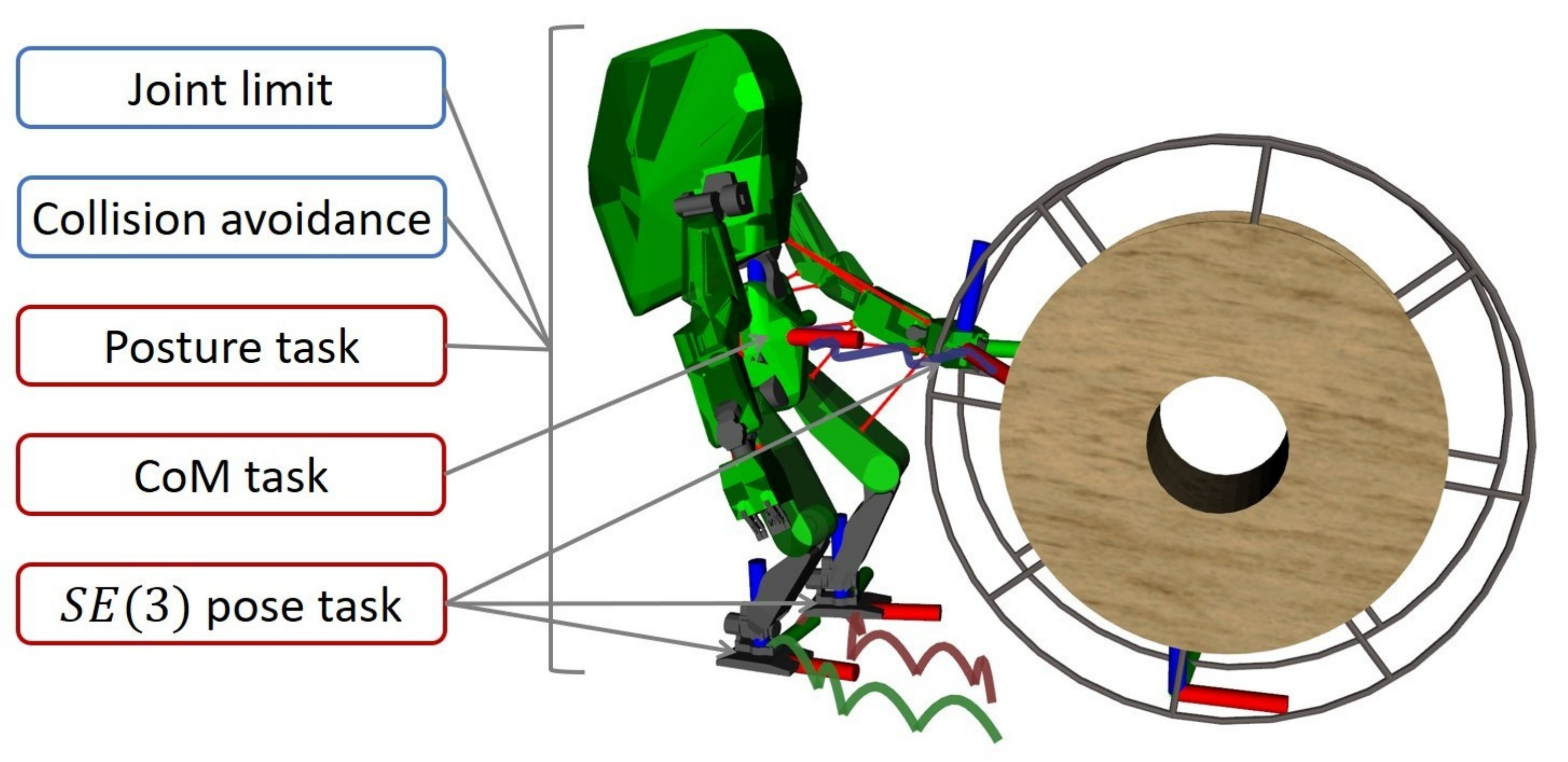}
    \caption{Tasks and constraints in WBM-planning.
      \newline \footnotesize
      Whole-body IK consists of three types of equality tasks (boxed in red)
      and two types of inequality constraints (boxed in blue).
      Links checked for self-collision are displayed in their green strictly convex hulls.
    }
    \label{fig:wbm-plan}
  \end{center}
\end{figure}

The trajectory of whole-body joint position is generated by the QP-based inverse kinematics (IK) calculation~\cite{PrioritizedIK:Kanoun:TRO2011}.
The target of the IK consists of three types of tasks as shown in \figref{fig:wbm-plan}: $\mathit{SE}(3)$ pose, CoM, and posture.
The $\mathit{SE}(3)$ pose tasks represent the target poses of the stance foot, swing foot, and hand grasping the object.
The swing foot trajectory is generated by interpolating the foot poses before and after stepping and the midpoint of the 50~mm height with a cubic spline curve.
In order to keep the torso link upright, the task of orientation in the roll and pitch directions is imposed with a small weight.
The target CoM is generated by the preview control~\cite{PreviewControl:Kajita:ICRA2003}
from the ZMP trajectory determined from the planned footstep sequence.
The task of half-sitting posture is imposed with a small weight for regularization.
The joint position limits and self-collision avoidance are formulated as inequality constraints.
The link shapes of the robot are represented by strictly convex hulls~\cite{SCH:Escande:TRO2014}.

\section{Application to Loco-manipulation Tasks} \label{sec:application}

The planning process was implemented in C++ and exchanges data via ROS communication~\cite{ROS:Morgan:ICRA2009WS}.
As external libraries,
OMPL~\cite{OMPL:Sucan:RAM2012} was used for RRT*~\cite{RRTstar:Karaman:IJRR2011} in OP-planning,
and SBPL~\cite{SBPL:github2020} was used for the AD* algorithm~\cite{ADstar:Likhachev:ICAPS2005} in FR-planning.

\subsection{Planning Results of Loco-manipulation Motions} \label{sec:planning-result}

We applied our proposed planning method to three types of loco-manipulation tasks.

\subsubsection{Bobbin Rolling Task} \label{sec:planning-result-cylindrical-object}

\begin{figure}[tpb]
  \begin{center}
    \includegraphics[width=1.0\columnwidth]{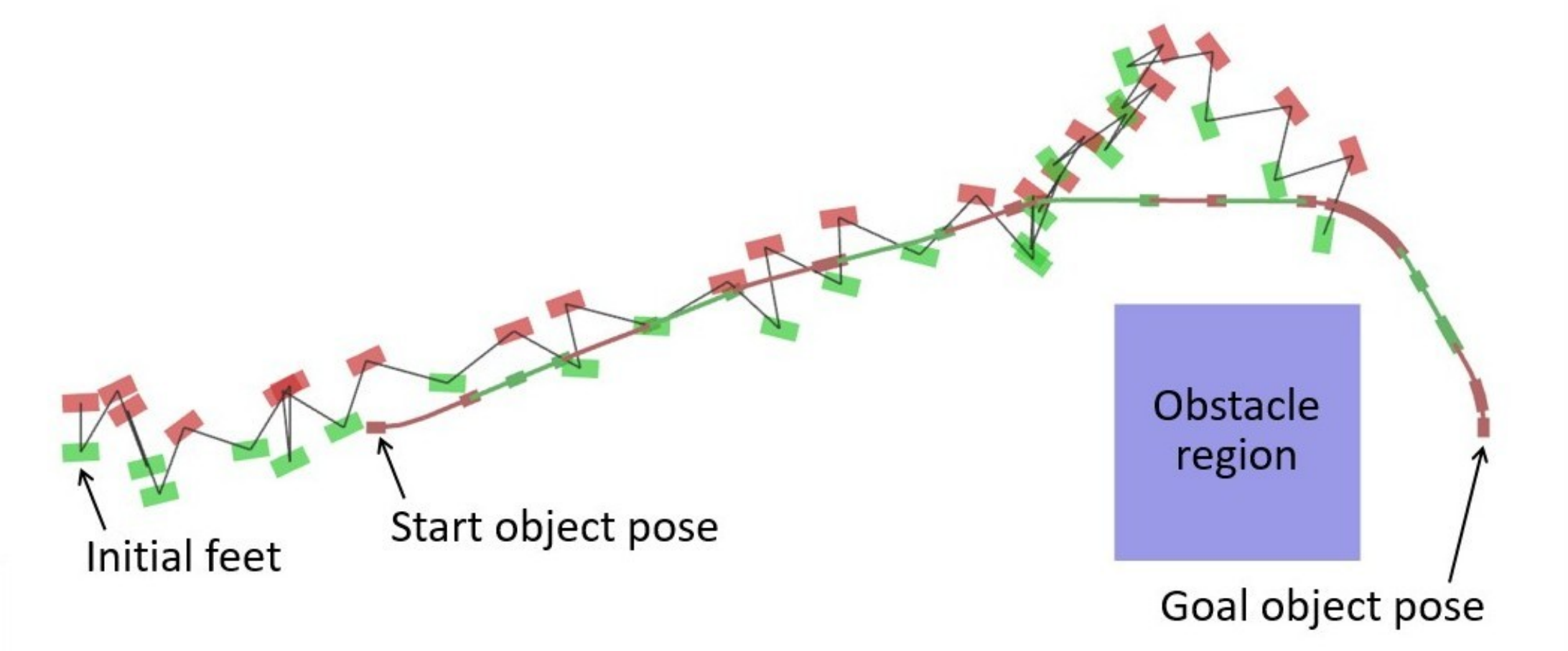}\\
    {\footnotesize (A) Result of OP-planning and FR-planning (top view)} \\
    \vspace{2mm}
    \includegraphics[width=1.0\columnwidth]{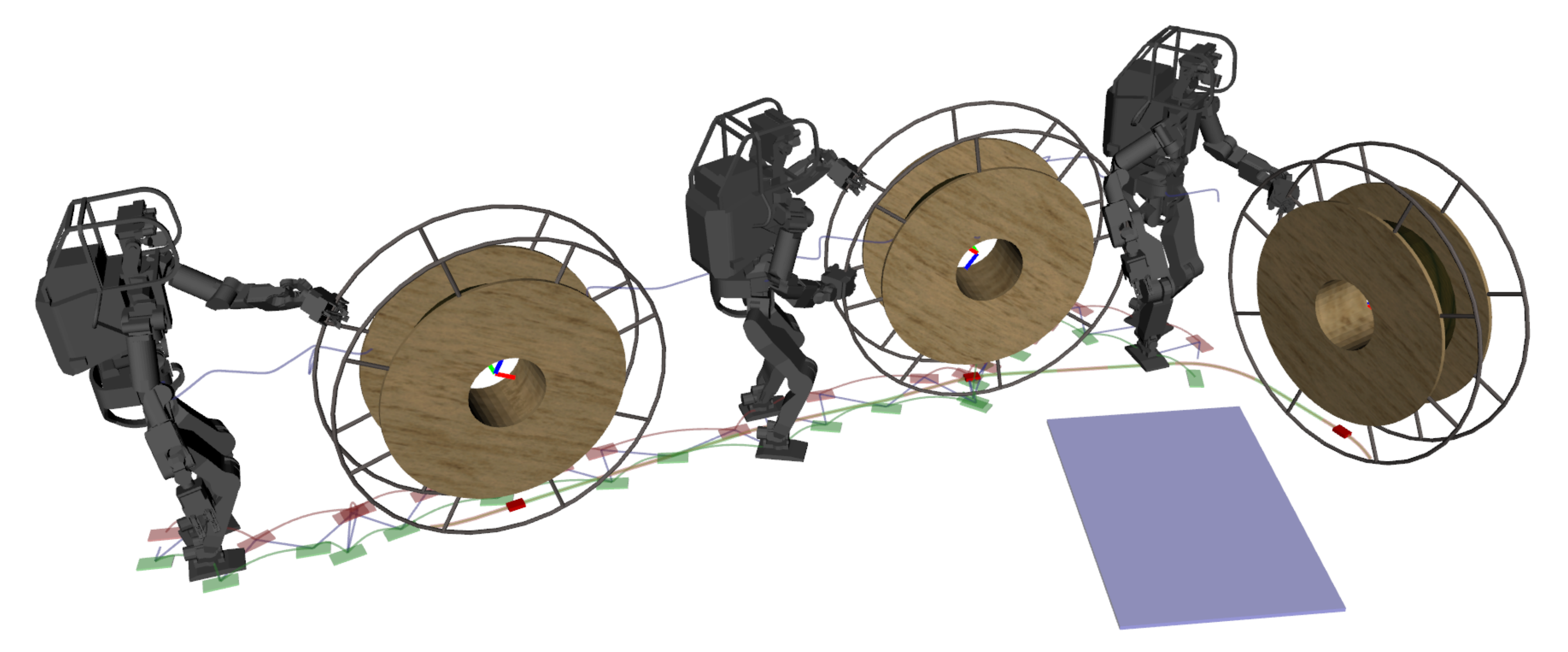}
    {\footnotesize (B) Result of WBM-planning}
    \caption{Bobbin rolling task.
      \newline \footnotesize
      HRP-5P humanoid carries a bobbin by rolling.
      The rectangle markers represent the footsteps of the left foot (red) and right foot (green).
      The sections of the object path that correspond to grasping with the left hand and right hand are drawn in red and green, respectively.
      The blue rectangle markers represent the obstacles.
    }
    \label{fig:drum-rolling}
  \end{center}
\end{figure}

\figref{fig:drum-rolling} shows the planning result of HRP-5P~\cite{HRP5P:Kaneko:RAL2019} moving-by-rolling a cable bobbin
with a handle of 1.5~m in diameter and 0.3~m in width.
Table~\ref{tab:planning-time} shows the computation time for each process.
A detailed analysis of FR-planning is provided later.

\begin{table}[h]
  \caption{Computation time of loco-manipulation planning}
  \label{tab:planning-time}
  \vspace{-3mm}
  \begin{center}
    \begin{tabular}{ccc|c}
      \hline
      OP-planning & FR-planning & WBM-planning & Total \\
      \hline
      0.51\,s (0.10\,s) & 1.15\,s (0.07\,s) & 3.90\,s & 6.19\,s \\
      \hline
    \end{tabular}
  \end{center}
  \begin{flushleft}
    The anytime algorithm is used for OP-planning and FR-planning,
    and the values in parentheses are the time until the initial solution is obtained.
  \end{flushleft}
  \vspace{-2mm}
\end{table}

\subsubsection{Door Opening Task}

\begin{figure}[tpb]
  \begin{center}
    \includegraphics[width=1.0\columnwidth]{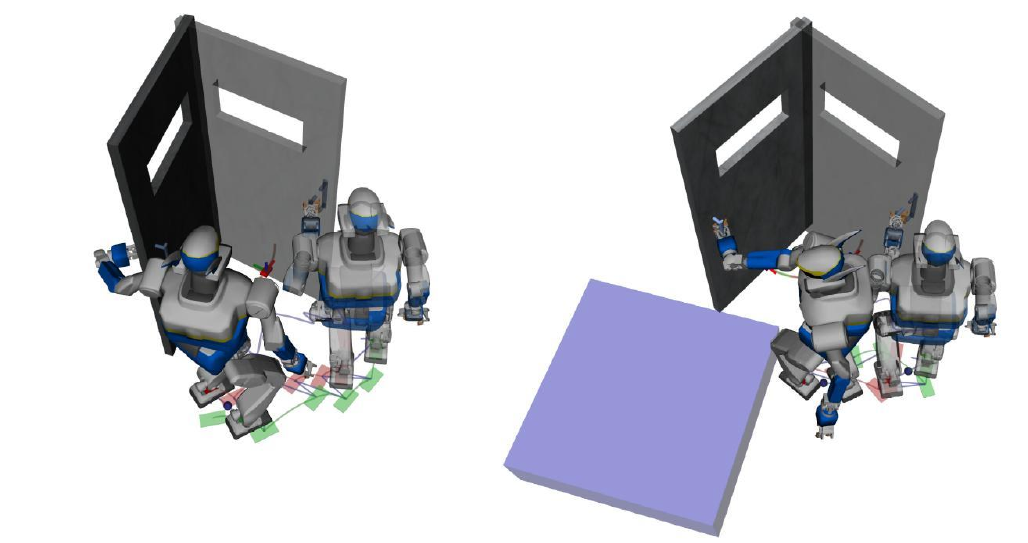}\\
    \begin{minipage}{0.49\columnwidth}
      \begin{center} \footnotesize (A) Without obstacle \end{center}
    \end{minipage}
    \begin{minipage}{0.49\columnwidth}
      \begin{center} \footnotesize (B) With obstacle \end{center}
    \end{minipage}
    \caption{Door opening task.
      \newline \footnotesize
      In (A), HRP-2Kai humanoid opens the door with the left hand until the end,
      whereas in (B), the robot switches from left to right hand to avoid the obstacle visualized in blue.
    }
    \label{fig:door-opening}
  \end{center}
\end{figure}

\figref{fig:door-opening} shows the planning result of HRP-2Kai opening a door.
Without obstacle, a motion without regrasping is planned with a small cost of the graph path.
Conversely, with an obstacle, the motion with regrasping from left to right hand is automatically planned to avoid the obstacle.
In this way, the proposed planner generates appropriate motions according to the environment.

\subsubsection{Cart Pushing Task}

\begin{figure}[tpb]
  \begin{center}
    \includegraphics[width=1.0\columnwidth]{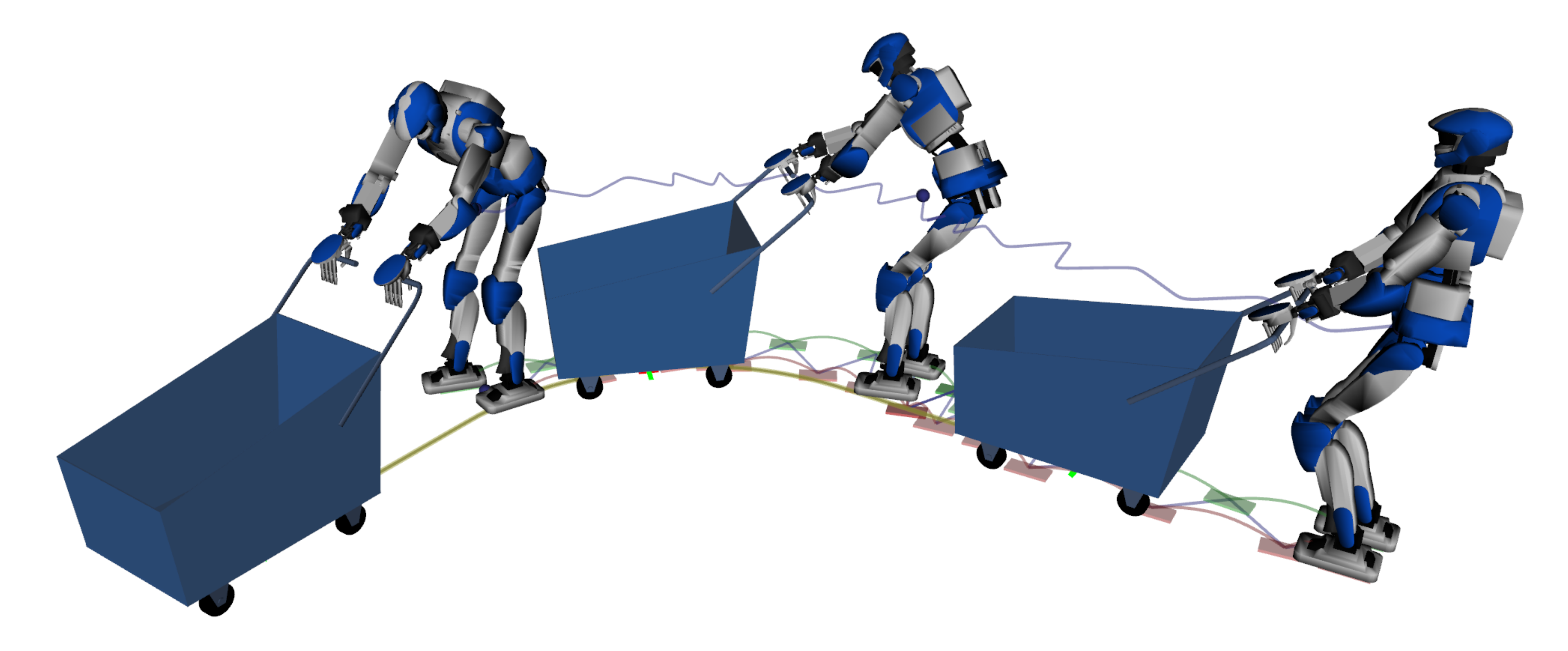}
    \caption{Cart pushing task.
      \newline \footnotesize
      HRP-4 humanoid carries the cart by pushing it with both hands.
    }
    \label{fig:cart-pushing}
  \end{center}
\end{figure}

\figref{fig:cart-pushing} shows the planning result of HRP-4 pushing a cart with both hands.
By setting $l_{\mathit{hand}}$ in state~\eqref{eq:state} to always be $L \land R$ and
using the intersection of the reachability maps of the left and right hands in the transition evaluation,
the proposed planner handles easily both hands grasp manipulations.

\subsection{Evaluation of Planning Results} \label{sec:result-evaluation}

\subsubsection{Computation Time}

Table~\ref{tab:fr-planning-result} shows the detailed results of FR-planning
in the examples described in Section~\ref{sec:planning-result}.
Compared with the state-of-the-art loco-manipulation planning~\cite{Locomanip:Jorgensen:ICRA2020},
it can be seen that our planner can quickly generate complicated and long-term motions.
In these planning examples,
sufficiently high-quality final solutions are obtained within the computation time limit set to a few seconds.
In addition, the initial solution is obtained in an extremely short time of approximately 100~ms;
this is an important feature for the robot to continue moving without stopping during planning.
The reachability maps are pre-generated with grid sizes of 100~mm and 10~degree; the computation time is approximately 300~s.
For the rolling objects, multiple reachability maps are generated for object-rolling angles from 0~degree to 45~degree in 5~degree increments.

\begin{table}[h]
  \caption{Detailed results of FR-planning}
  \label{tab:fr-planning-result}
  \vspace{-3mm}
  \begin{center}
    \begin{tabular}{cc||ccccc}
      \hline
      & & (a) & (b) & (c) & (d) & (e) \\
      \hline
      \multirow{2}{*}{Bobbin} &
      Initial sol. & 0.07\,s & 82.00 & 11.63 & 51 & 6524 \\
      & Final sol. & 1.15\,s & 3.60 & 11.02 & 45 & 45673 \\
      \hline
      Door
      & Initial sol. & 0.04\,s & 40.20 & 3.02 & 13 & 2634 \\
      w/o obstacle
      & Final sol. & 0.47\,s & 3.60 & 2.01 & 13 & 21727 \\
      \hline
      Door
      & Initial sol. & 0.04\,s & 75.80 & 3.02 & 13 & 2634 \\
      w/ obstacle
      & Final sol. & 0.05\,s & 7.20 & 3.02 & 13 & 2669 \\
      \hline
      \multirow{2}{*}{Cart} &
      Initial sol. & 0.13\,s & 81.20 & 7.58 & 41 & 9849 \\
      & Final sol. & 1.88\,s & 1.80 & 6.18 & 29 & 74991 \\
      \hline
    \end{tabular}
  \end{center}
  \begin{flushleft}
    FR-planning performance for Section~\ref{sec:planning-result} use-cases:
    (a) computation time,
    (b) heuristics inflation factor~\cite{ADstar:Likhachev:ICAPS2005},
    (c) cost from start to goal,
    (d) length of the planned footstep sequence,
    (e) number of expanded states.
    The heuristics inflation factor qualifies the solution optimality: the closer it is to~1, the more optimal the solution is.
    The use-case of door with obstacle only expands a few more states in the final solution than the initial solution
    because the initial solution is already high-quality comparable to the final solution.
  \end{flushleft}
  \vspace{-2mm}
\end{table}

\subsubsection{Footstep Actions}

\begin{figure}[tpb]
  \begin{center}
    \includegraphics[width=0.335\columnwidth]{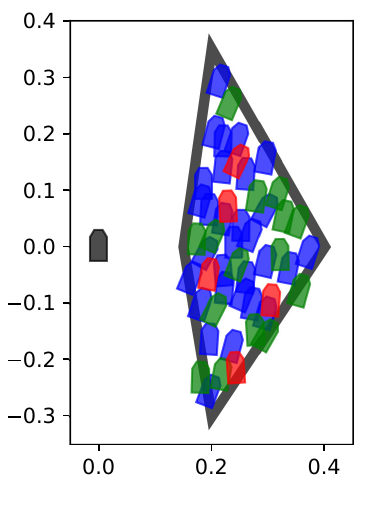}
    \includegraphics[width=0.65\columnwidth]{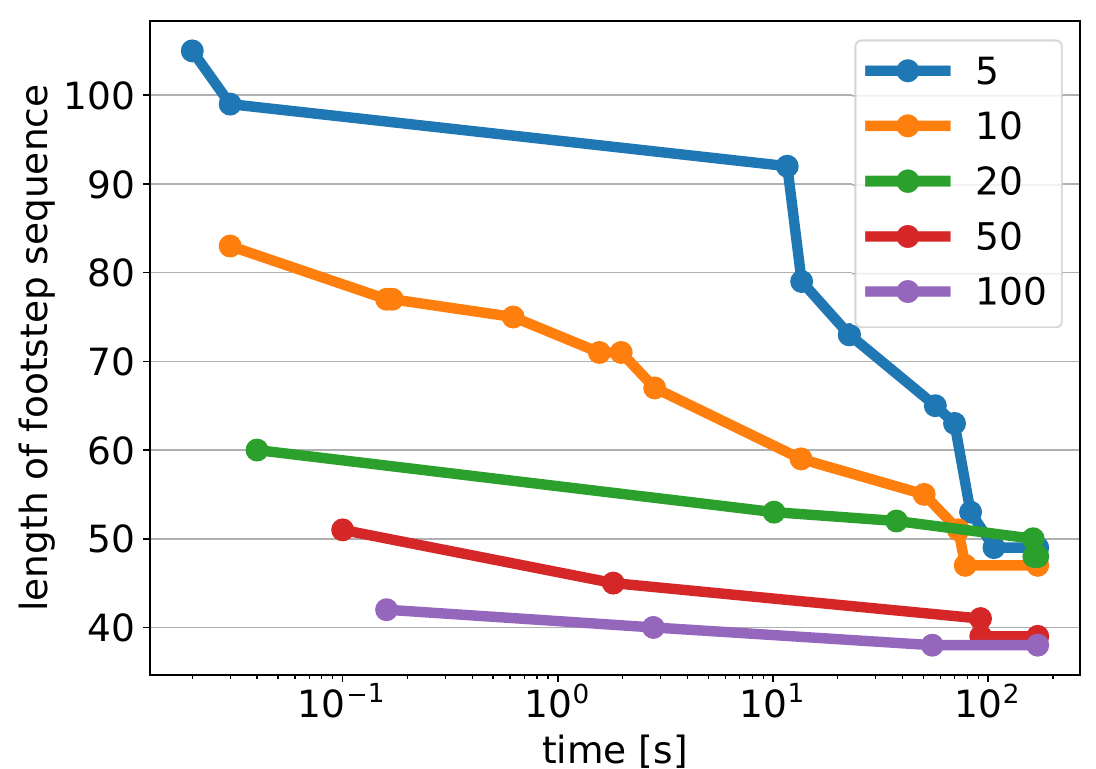} \\
    \begin{minipage}{0.335\columnwidth}
      \begin{center} \footnotesize (A) Footstep action set \end{center}
    \end{minipage}
    \begin{minipage}{0.65\columnwidth}
      \begin{center} \footnotesize (B) Computation time and solution quality \end{center}
    \end{minipage}
    \caption{Footstep actions in FR-planning.
      \newline \footnotesize
      (A) The black foot marker represents the stance foot,
      and the area surrounded by the black line represents the allowable area for the landing position of the swing foot.
      The red, green and blue foot markers represent the footstep actions generated by the Halton sequence
      (red: $1,\cdots,5$; green: $6,\cdots,20$; blue: $21,\cdots,50$).
      \newline \footnotesize
      (B) The transition of the length of the planned footstep sequence is shown
      for the number of actions $N_\mathcal{A}$ of 5, 10, 20, 50, and 100,
      The smaller the length of the footstep sequence, the higher the quality of the solution.
      For large number of actions, it takes time to obtain the initial solution, yet its quality is high.
    }
    \label{fig:footstep-actions}
  \end{center}
\end{figure}

As mentioned in Section~\ref{sec:fr-planning-settings}, in FR-planning, footstep action set $\mathcal{A}$ is prepared in advance.
We use the Halton sequence~\cite{HaltonSeq:Halton:NM1960}, which has deterministic and quasirandom properties,
to generate the footstep action set (\figref{fig:footstep-actions}~(A)).
\figref{fig:footstep-actions}~(B) shows the relationship
between the number of actions $N_\mathcal{A}$ and the length of the footstep sequence planned
in Section~\ref{sec:planning-result-cylindrical-object}.
The plan is robust regardless of the number of actions;
however, it can be seen that a high-quality initial solution is obtained when the number of actions is large.

\subsubsection{Heuristic Function}

We evaluate the effectiveness of $h_{\mathit{nominal}}$ in the heuristic function~\eqref{eq:heuristics}
for the planning use-cases of Section~\ref{sec:planning-result-cylindrical-object}.
The nominal pose of the foot is set to the pose 1.2~m behind the current object.
Table~\ref{tab:heuristic-time} shows a comparison of the computation time
until the initial solution is obtained with and without $h_{\mathit{nominal}}$.
Especially for difficult problems such as moving a large object along a curved path,
this heuristic avoids increasing the computation time by preferentially expanding the states leading to the goal.

\begin{table}[h]
  \caption{Computation time with and without heuristics}
  \label{tab:heuristic-time}
  \vspace{-3mm}
  \begin{center}
    \begin{tabular}{c||cc}
      \hline
      & straight path & curved path \\
      \hline
      w/ heuristics of nominal pose & 0.03\,s & 0.07\,s \\
      \hline
      w/o heuristics of nominal pose & 0.43\,s & 17.41\,s \\
      \hline
    \end{tabular}
  \end{center}
  \begin{flushleft}
    The time until the initial solution is obtained is shown.
    FR-planning was performed on straight and curved object paths.
  \end{flushleft}
  \vspace{-2mm}
\end{table}

\subsubsection{ZMP Evaluation}

\begin{figure}[tpb]
  \begin{center}
    \includegraphics[width=1.0\columnwidth]{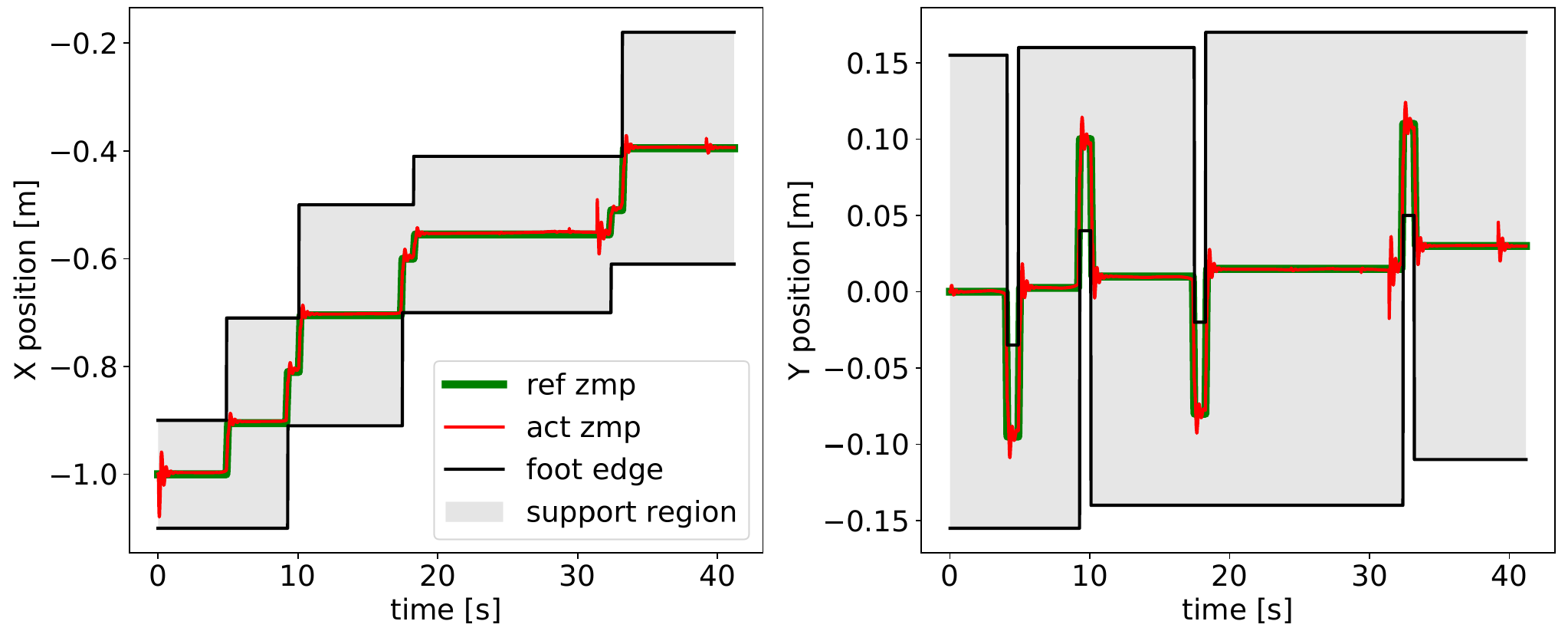}
    \caption{ZMP trajectory of the planned motion.
    }
    \label{fig:zmp-result}
  \end{center}
\end{figure}

Although the robot motions in Section~\ref{sec:planning-result} were only animated
with the visualization software Rviz~\cite{ROS:Morgan:ICRA2009WS},
we checked the dynamics validity of the motion based on ZMP.
\figref{fig:zmp-result} shows the ZMP trajectory
when the robot moves forward several steps while rolling the bobbin.
The planned motion is dynamically feasible as the ZMP is within the support region.

\section{Conclusion}

In this letter,
we proposed a versatile planning framework for loco-manipulation.
In order to quickly and flexibly generate complex loco-manipulation motion involving object regrasping and obstacle avoidance,
we introduced a transition model that can be evaluated efficiently based on reachability maps
in a sophisticated graph search algorithm.
We have shown the effectiveness of our planning framework
by applying it to various loco-manipulation tasks
such as rolling an object by a humanoid robot.
Future challenges include
improving parallelization of locomotion and manipulation,
consideration of reaction forces from the object, and
real-time replanning based on object tracking.






\bibliographystyle{IEEEtran}
\bibliography{main.bib}

\end{document}